%% file: main.tex
\documentclass[letterpaper,twocolumn,10pt]{article}
\usepackage{usenix}
\usepackage{colortbl}
\usepackage{textcomp}
\usepackage{graphicx}
\usepackage{makecell}
\usepackage[ruled,vlined,linesnumbered]{algorithm2e}
\SetAlgoSkip{}
\graphicspath{ {./images/} }

\usepackage{amsmath}
\usepackage{url}
\usepackage{caption}
\usepackage{subcaption}
\usepackage{booktabs}
\usepackage{multirow}
\usepackage{enumitem}

\usepackage{xcolor, color, soul}
\let\oldnl\nl
\newcommand{\nonl}{\renewcommand{\nl}{\let\nl\oldnl}} 

\let\mc\mathcal

\begin{document}

 \author{\rm Trishita Tiwari \\
 Cornell University
\and
 \rm Suchin Gururangan \\
 University of Washington
\and
 \rm Chuan Guo \\
 FAIR at Meta 
\and
 \rm Weizhe Hua \thanks{Work done while at Cornell University.}\\
 Google DeepMind
\and
 \rm Sanjay Kariyappa \\
 Georgia Institute of Technology
\and
 \rm Udit Gupta \\ Cornell University
\and
 \rm Wenjie Xiong \\ Virginia Tech
\and
 \rm Kiwan Maeng \\ Pennsylvania State University
\and
 \rm Hsien-Hsin S. Lee \thanks{Work done while at Meta.} \\ Intel 
\and
 \rm G. Edward Suh 
 \textsuperscript{\textcolor{green}{$\dagger$}} \\ NVIDIA / Cornell University
 }

\title{Information Flow Control in Machine Learning \\through Modular Model Architecture}

\maketitle

\input{abstract}
\input{introduction2}

\input{ifc-ml}

\input{our-solution}
\input{experiments}
\input{ablation}
\input{limitations}
\input{extension}
\input{related-works}
\input{conclusion}

\bibliographystyle{abbrv}
\bibliography{refs}

\appendix

\input{appendix}

\end{document}

%% file: abstract.tex
\begin{abstract}

In today's machine learning (ML) models, any part of the training data can affect the model output.
This lack of control for information flow from training data to model output is a major obstacle in training models on sensitive data when access control only allows individual users to access a subset of data.  To enable secure machine learning for access-controlled data, we propose the notion of information flow control for machine learning, and develop an extension to the Transformer language model architecture that strictly adheres to the IFC definition we propose. Our architecture controls information flow by limiting the influence of training data from each security domain to a single expert module, and only enables a subset of experts at inference time based on the access control policy.
The evaluation using large text and code datasets show that our proposed parametric IFC architecture has minimal (1.9\%) performance overhead and can significantly improve model accuracy (by 38\% for the text dataset, and between 44\%--62\% for the code datasets) by enabling training on access-controlled data.

\end{abstract}

%% file: introduction2.tex
\vspace{-0.1in}
\section{Introduction}
\label{sec:intro}
\vspace{-0.1in}

Recent studies~\cite{carlini2021extracting, carlini_chatgpt, carlini_diffusion} showed that large machine learning (ML) models can leak sensitive information that was in their training data through their inference-time output.
This inference-time leakage introduces a new security/privacy concern when different users have access to different (training) data.
%
For example, consider a scenario where a company trains an in-house code-completion model~\cite{copilotapi} with its proprietary code repositories, when individual employees have access to different subsets of the repositories based on their team or projects.
The company faces a dilemma---if the model is trained with all the repositories, employees may be able to learn about the code that they do not have permission to access; if only trained with repositories that everybody has access to, the model quality will degrade.
It would be ideal if the model can generate different outputs to different employees, by selectively leveraging training data that they have access to. 
Unfortunately, 
today's ML models mix information from all training data during the training process, and cannot selectively utilize a subset of training data.

%

To address this problem, we introduce the notion of information flow control (IFC) to machine learning. 
Instead of treating all training data uniformly, we partition the training dataset into multiple \emph{security domains}. During inference, a trained ML model takes an \emph{access policy} that represents which security domains are accessible by the current user and ensures its output does not leak any training data that the user cannot access. 
The goal of IFC in this ML setting is to develop a training process, an inference process, and a model architecture that can honor the access policy at inference time by guaranteeing no leakage of information from inaccessible training data to the model output. 
We define the problem through the lens of non-interference (NI), and present IFC as a new challenge that ML systems should address. 

%
%

Deep learning models are notoriously difficult to understand or control even though they perform well in practice. In that sense, IFC in parametric deep neural networks may seem infeasible at a glance.
We show that information flow can be controlled at the model architecture level.
Our approach trains a separate, small sub-module, called an \emph{expert} for each security domain, so that each security domain's data only influences a single expert.
Then, we develop a secure \emph{gating} function that is used at run-time to activate only a small number ($k$) of the most relevant, accessible experts for each user query considering the user's access policy. 
Once the gating function picks the top-$k$ experts, the final output is generated by \emph{aggregating} the outputs/parameters of the $k$ experts.
To ensure non-interference (NI), the gating and aggregation decisions only use information from accessible security domains.
%

While people studied modular model architectures~\cite{hu2021lora, masoudnia2014mixture, gururangan2021demix} for non-security purposes, 
we found that designing a secure modular architecture for IFC introduces a new set of technical challenges.
For IFC, in addition to only using accessible experts at inference time, the gating decisions (choosing which experts to activate) and aggregation methods (combining information from the different activated experts) must also satisfy non-interference. 
This condition implies that gating and aggregation functions cannot be learned from the entire training data, as is the case with many of today's ensemble or mixture-of-experts (MoE) models. 
Moreover, the access policy is only given at run-time, and the decisions must be made quickly with only information from the accessible domains, which may change for each query. 
%
Additionally, the gating function must scale to a large number of security domains, which may be dynamically added or removed over time.
We address these challenges by designing three new secure gating functions, and two secure expert aggregation methods based on output ensembling and parameter merging. 
The secure gating functions allow our IFC architecture to handle both cases when the target domain of an input is known and unknown, and a large number of security domains.  



We implement our IFC design by extending two Transformer language models (GPT-$2$ and OPT) with our secure architecture, and evaluate them with real-world text~\cite{baumgartner2020pushshift} and code~\cite{codeparrot} data with simulated security domains.
The evaluation results show that our proposed architecture can improve the accuracy (perplexity) by 38--62\%, while strictly following the given access policy.
When multiple experts are evaluated in parallel, the execution time of the secure IFC model is also close to the insecure baseline, adding at most 1.9\% latency.
The proposed parametric IFC architecture can also be useful for attribution, as only a small subset of activated experts influence the output, and machine unlearning, as it allows an easy removal of a security domain.

%
%
In summary, our technical contributions are as follows:
\begin{enumerate}[noitemsep, leftmargin=*, topsep=0pt]
    \item We introduce information flow control for ML and formally define non-interference in the ML context.
    \item We propose modifications to the general Transformer architecture to enable information flow control. We propose fast, effective, and scalable gating functions and expert aggregation methods to enable information from multiple accessible security domains to be efficiently combined at run-time, improving the output quality while strictly enforcing non-interference from inaccessible domains. 
    \item We evaluate our approach with large corpuses of text~\cite{baumgartner2020pushshift} and code~\cite{codeparrot} datasets and show that we can enable secure use of access-controlled data with low performance overhead. The ability to securely use in-domain training data can significantly improve model accuracy compared to the baseline that is only trained on public data.
\end{enumerate}





%% file: ifc-ml.tex
\vspace{-0.25in}
\section{Information Flow Control (IFC) in ML}
\label{sec:ifc-ml}
\vspace{-0.1in}



Here, we discuss information flow control (IFC) for machine learning (ML), including the motivation, a formal definition, and high-level approaches to realize it.
The next section introduces a concrete realization of IFC for language models.


\vspace{-0.1in}
\subsection{Objective and Setup}
\vspace{-0.05in}

\begin{figure}
\begin{center}
\includegraphics[trim={0.1cm 0.15cm 0.3cm 0.2cm},clip]{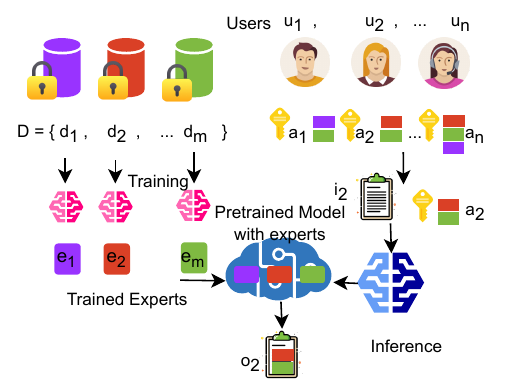}
\vspace{-0.15in}
\caption{\small Information flow control (IFC) in machine learning. 
}
\label{fig:problem-statement}
\vspace{-0.31in}
\end{center}
\end{figure}

The goal of IFC in ML is to ensure that a machine learning output only depends on a subset of training data that a user has access to.
Figure~\ref{fig:problem-statement} illustrates the setup and how IFC changes training and inference for ML. 
We assume a setting where the training data set $D$ is partitioned into $m$ different {\em security domains}, i.e., $D = \{d_1, d_2$ ... $d_m\}$.
There are $n$ users ($u_1$ ... $u_n$), each with an \emph{access policy} ($a_1$ ... $a_n$), which defines which subset of security domains out of $D$ are accessible to each user. At inference time, a user (say user $u_2$) provides an input ($i_2$) and an access policy ($a_2$) to the model. Then, the model with IFC capability needs to ensure that the corresponding output ($o_2$) depends only on the domains from $D$ that are present in the access policy $a_2$. 

\vspace{-0.1in}
\subsection{Example Use Cases for IFC in ML}
\label{sec:use-cases}
\vspace{-0.05in}

The ability to control information flow from training data to model outputs is important in many use cases where different users can only access different subsets of the training data. Below, we briefly discuss example use cases.

\noindent {\bf AI Coding Assistant.}
In order to provide more relevant suggestions, a code completion model for a coding assistant can be trained using code from a company's internal repositories.
However, code repositories typically have access control, and individual engineers only have access to a subset of repositories.
A code completion model needs to honor the access restrictions and only suggest code based on the repositories that a particular user has access to at inference time.   
The security domains for code repositories can be delegated based on projects or organizations. 
In this paper, we study this case by treating a Github repository as a security domain so that it can have its own access permission. 

%

\noindent {\bf AI for Writing/Summarizing Documents.}
A language model (LM) can be used to provide text suggestions when writing or summarizing documents with sensitive content, such as company internal memos, project reports, military documents, etc. 
The language model can perform better if it can leverage related (sensitive) documents that the user has access to.
For instance, when writing a project report, a language model accessing other documents from the same project will provide more tailored suggestions. 
However, we need to ensure that the model's suggestions do not leak any information from other inaccessible documents.
For example, access to company documents may be limited based on projects or organizations, and
individual users may have access to a different set of projects or organizations.
In this case, the documents from the same project/organization can be put into a separate security domain.
We use the Pushshift.io dataset to simulate such use cases.

\noindent {\bf AI-based Q/A on Access Controlled Materials.}
The recent advances in large language models (LLMs) showed a potential for use in search or Q/A engines (e.g., ChatGPT, Bing Chat). 
While these models are only trained on public domain data today, it is likely that the LLM-based search or Q/A engines will also be applied to answer questions based on access controlled data such as pay-walled content (e.g., Wall Street Journal, New York Times, etc), internal corporate documents, etc.  
In such cases, the LLM may need to incorporate the access controlled data for the best performance.
However, in order to prevent unintended leakage of access controlled data, the model should be able to ensure that its output only depends on the subset of the training data that a particular user is allowed to access.

\vspace{-0.1in}
\subsection{Extending IFC Definition to ML}
\label{sec:ni}
\vspace{-0.05in}

We define IFC in ML by extending the notion of non-interference (NI) in the traditional information flow control.

\noindent {\bf Traditional Non-Interference (NI).}
Non-Interference~\cite{goguen1982security} is a security policy commonly used in information flow control to limit the influence between security levels. Let $D$ represent the machine state, and let $d_L$ and $d_H$ be the projection of the machine state $D$ to the low sensitivity and high sensitivity parts, respectively. Let $=_L$ be the function that compares the low sensitivity parts of the state, i.e., $D =_L D^{\prime}$ iff $d_L = d_L^{\prime}$. Let $(P, D_1) \Downarrow D_2$ be the execution of a program $P$ starting with machine state $D_1$ and terminating with the machine state $D_2$. Given this, Non-Interference for a deterministic program $P$ is defined as follows~\cite{goguen1982security}:
\begin{align*}
    \forall D_1, D_1^{\prime} : D_1 &=_L D_1^{\prime} \wedge \\
    (P, D_1) &\Downarrow D_2 \wedge \\
    (P, D_1^{\prime}) &\Downarrow D_2^{\prime} \Rightarrow \\
    D_2 &=_L D_2^{\prime}
\end{align*}
In other words, a deterministic program $P$ satisfies Non-Interference if, given the same initial low sensitivity machine state, its transformation of the low sensitivity machine state is the same regardless of the high sensitivity machine state.
\noindent {\bf Non-Interference for Machine Learning.} 
%
%
%
%
We define non-interference in machine learning below. A generalized definition  for probabilistic protection is presented in Section~\ref{sec:extension}.

\textbf{\em Definition} Let $D = \{d_1$ ... $d_m\}$ be a dataset partitioned into $m$ security domains.
For a given user $u_i$, we define the user's \emph{access policy} $a_i \subseteq \{1,\ldots,m\}$, which specifies a set of domains accessible by $u_i$. For two datasets $D=\{d_1,\ldots,d_m\}$ and $D^{\prime}=\{d_1^{\prime},\ldots,d_m^{\prime}\}$, let $=_{a_i}$ be the comparison operator such that $D =_{a_i} D^{\prime}$ iff $d_j = d_j^{\prime}$ for all $j \in a_i$.
%
Let $\mc{M}_D(a_i, \pmb{x}) \Downarrow o$ denote the inference algorithm for a model $\mc{M}$ that was trained on dataset $D$, which performs inference for request $\pmb{x}$ with access policy~$a_i$ and the inference result being a distribution $o$ over all possible outputs.
We say that $\mc{M}$ satisfies Non-Interference (NI) with respect to access policy $a_i$ if for all queries $\pmb{x}$ the following holds:
\begin{align*}
    \forall D, D^{\prime}: D =_{a_i} D^{\prime} \wedge \\
    \mc{M}_D(a_i, \pmb{x}) &\Downarrow o \wedge \\
    \mc{M}_{D^{\prime}}(a_i, \pmb{x}) &\Downarrow o^{\prime} \Rightarrow \\
    o &= o^{\prime} 
\end{align*}
This definition prohibits protected data from influencing the output (or their likelihoods, when the output is probabilistic).
Our definition is also \emph{termination sensitive}, meaning that the termination behavior (when the output generation will be finished) is not influenced by the protected data as well.

\vspace{-0.1in}
\subsection{Parametric and Non-Parametric IFC}
\label{sec:rag}
\vspace{-0.05in}


The main technical challenge in IFC for ML comes from the difficulty in controlling or tracking the information flow through the trained model parameters. 
One approach to sidestep this challenge is to only use non-parametric methods to leverage sensitive data: instead of using sensitive data for training model parameters, the data is protected using the traditional access control or IFC mechanisms and only retrieved and incorporated to an ML model at inference time.
For example, retrieval augmented generation (RAG) \cite{guu2020retrieval,lewis2020retrieval,borgeaud2022improving,khandelwal2019generalization} showed that incorporating retrieved data as a part of the input context of a language model is an effective way to complement a parametric language model (LM), especially for knowledge-intensive tasks such as question answering.

Parametric training (fine-tuning) and non-parametric RAG are complementary in their nature and we believe that both will be important in achieving the state-of-the-art AI/ML capabilities going forward. 
In that sense, we need IFC for ML to be able to handle both parametric training and non-parametric retrieval augmentation. 
As non-parametric IFC can largely leverage existing access control and IFC techniques, in this paper, we focus on enabling IFC for parametric models.

The following discussion further elaborates why we believe that both parametric and non-parametric IFC will be necessary to fully utilize sensitive data in ML.

\noindent {\bf Complementary Capabilities.}
Today, RAG is primarily considered as a way to further improve LMs in addition to fine-tuning rather than replacing fine-tuning.
RAG allows more recent information to be added/updated without re-training, improving attribution and enabling faster adaptation/customization.
On the other hand, fine-tuning is generally considered to be necessary for deeper adaptation of LLM behaviors such as changing writing style/tones and adding new domain-specific knowledge, terminologies, etc. 
Today's RAG approaches also increase the context length of LMs and thus inference costs, especially when knowledge from multiple sources is needed to be assembled \cite{xu2023recomp,fan2019eli5}.
These costs fundamentally limit the amount of new information that can be incorporated at inference time in practice.
Intuitively, adding a small amount of information to the context can work well for some tasks such as generating answers to specific questions. Yet, fine-tuning will likely be more effective in learning from a large amount of data, merging knowledge from multiple domains, etc.

\noindent {\bf Need for Parametric Training in RAG.}
Studies in RAG found that off-the-shelf parametric models often struggle to leverage additional retrieved information at inference time \cite{shi2023large} and parametric fine-tuning is still needed.
For more effective RAG models, the encoder model that generates embedding vectors for search and the decoder model that generates outputs need to be trained/fine-tuned \cite{izacard2022few,guu2020retrieval}, making our work still relevant. 

\noindent {\bf Models w/o RAG.}
While generative models can often benefit from retrieved data at inference time, many ML models will likely remain mostly parametric. 
For example, it is not clear how retrieval augmentation can be used to allow image classification models to leverage classified images.
For such parametric models, we need the parametric IFC capability.

\begin{figure} 
\begin{center}
\includegraphics[]{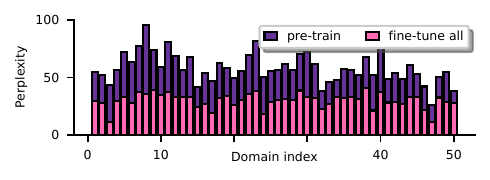}
\vspace{-0.2in}
\caption{\small Pre-trained GPT-2 evaluated on 50 datasets from Pushshift.io consistently underperforms fine-tuned GPT-2. \label{fig:baseline}}
\end{center}
\vspace{-0.35in}
\end{figure}

\vspace{-0.1in}
\subsection{Comparison with Differential Privacy}
\label{sec:dp}
\vspace{-0.05in}

Differential Privacy (DP)~\cite{dwork2006calibrating} is another method to ensure privatization of sensitive data. A randomized algorithm is said to be differentially private if it produces similar output regardless of whether the sensitive data was included in the algorithm's database.
Unlike IFC which allows users with different access policies to selectively benefit from different subsets of security domains within the protected data, DP limits all users from benefiting from all protected data. Moreover, IFC achieves zero leakage, unlike DP which still leaks data (captured by its privacy parameters, e.g., $\epsilon$).

Furthermore, DP typically aims to protect individual data samples (i.e., example-level DP), and the privacy protection becomes weaker when a group of samples are correlated and must be protected together~\cite{kifer_no_free_lunch} (e.g. multiple texts in a security domain). Applying DP to protect a group of data requires group-level DP~\cite{dwork2014algorithmic}, which amplifies the privacy parameter $\epsilon$ by the size of the group and thus hurts utility proportional to the size of the group. 
Recent studies on DP-finetuning of GPT-2~\cite{li2021large,bu2024automatic,shi2022just} have shown noticeable accuracy loss of between 1\% and 40\%, even with example-level DP, at a moderate privacy parameter ($\epsilon=8$).
The utility loss will be drastically amplified when a group of data in each security domain must be protected together, as our security domains are large (3K--900K samples per domain in Pushshift dataset). 

Nevertheless, we note that DP may be more applicable than IFC in certain scenarios. 
For example, if there are a very large number of security domains, say one domain per each user for millions or billions of users, user-level DP can be a natural way to guarantee data privacy with moderate utility loss. 
In comparison, achieving IFC using our approach would require training one adapter per user, which can be prohibitively expensive to scale.

\vspace{-0.1in}
\subsection{Limitations of Naive IFC Solutions}
\vspace{-0.05in}

Here, we discuss a few simple approaches to leverage sensitive training data with access restrictions in parametric models, and discuss why these coarse-grained approaches available today are unlikely to be sufficient in practice.

\noindent {\bf Train Only on Public Data.} One simple option is to use a model trained only on public data, which intuitively cannot leak any sensitive data.
However, models trained only on public data cannot perform specific tasks well compared to fine-tuned models. Figure~\ref{fig:baseline} shows the perplexity (the measure of accuracy; lower is better) of GPT-2 on data from $50$ different subreddits (security domains) from Pushshift.io. We see that the model fine-tuned on (potentially sensitive) additional data significantly outperforms the same model trained with only public data.


\noindent {\bf Fine-tune a Model for Each Access Policy.} 
We can fine-tune a new model for each possible access policy. For $m$ domains, this requires fine-tuning at most $2^m$ different models. The exponential growth makes this approach impractical unless $m$ is very small.

\noindent {\bf Fine-tune One Model for Each Security Domain.} 
Another option is to fine-tune one model for each security domain (only train $m$ models), and choose one of the models that a user has access to at run-time. 
However, our experiment later (Section~\ref{sec:ablation}) shows that this approach is sub-optimal, and a user with access to multiple domains can significantly improve the model's output quality by using multiple domains at the same time. 
\vspace{-0.1in}
\subsection{Our Approach: IFC through Modular Architecture}
\label{sec:operationalization}
\vspace{-0.05in}

\begin{figure}
    \centering\includegraphics{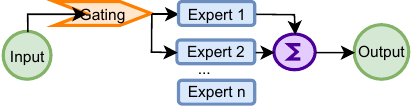}
\vspace{-0.1in}
    \caption{\small Parametric IFC through modular model architecture. 
    The influence of each security domain is limited to one expert.
    A \emph{gating function}, at inference time, decides which of the experts should be activated for a given input. 
    The activated experts are aggregated either though ensembling their outputs or aggregating the parameters.
    }
    \label{fig:moe}
\vspace{-0.2in}
\end{figure}

At a high-level, IFC for a parametric model needs the following three components:

\begin{enumerate}[noitemsep, leftmargin=*, topsep=0pt]
    \item 
    \textbf{Parameter Isolation.} During training, there must be a way to isolate parameters influenced by different security domains. We refer to these isolated parameters as \emph{experts}. 
    
    \item 
    \textbf{Secure Gating.} With a large number of security domains, it is infeasible and ineffective to use all accessible domains. 
    At inference, a secure gating function needs to determine which of the accessible domains are relevant to the current input in a way that does not use information from the inaccessible domains. We refer to this process as \emph{gating}.
    
    \item 
    \textbf{Secure Expert Aggregation.} Finally, the information from the domains chosen during gating needs to be combined to produce one output. This act of combining information from the different chosen domains can also only use information from accessible domains. We refer to this step as \emph{aggregation}.
\end{enumerate}

To enable more fine-grained parametric IFC, we propose an approach based on a modular model architecture as shown in Figure~\ref{fig:moe}. 
In this approach, we separately train a different expert with data from each security domain to ensure parametric isolation. Then, at run-time, once we have the user's input and access policy, we use a secure gating algorithm to find the top-$k$ best-suited experts permitted by the access policy.
Once we have our top-$k$ experts, we use them in our aggregation scheme to produce one final output for the user. Each of these steps is carefully designed to ensure no information is leaked from domains that are not allowed by the access policy. 

While there have been previous studies on modular ML model architectures~\cite{hu2021lora, masoudnia2014mixture, gururangan2021demix}, the previous designs largely investigated modularity in the context of improving inference efficiency. 
We found that designing a modular architecture for IFC introduces a new set of technical challenges.

\noindent {\bf Secure Real-time Gating.}
IFC presents unique challenges to gating that have not been explored before. 
In traditional modular architectures such as Mixture of Experts (MoE)~\cite{masoudnia2014mixture}, gating functions used to determine expert relevance are often learned during model training using information from the entire training set. 
Such learned gating functions cannot enforce isolation of inaccessible experts and cannot be used for IFC.
This is because information from training data can leak through which experts are activated and how the experts are aggregated even if only accessible experts are activated at run-time.
Moreover, because the access policy is only available at run-time, gating decisions in IFC cannot heavily rely on offline computation and thus need to quickly rank accessible domains at run-time only using information from said accessible domains.
In all, we must now construct a real-time gating scheme that is fast, incorporates domain access restrictions, scales to a large number of domains, and maps the user's limited text input to relevant domains. 
We address this challenge for LMs by developing three different secure real-time gating schemes in Section~\ref{sec:gate-k}. 


\noindent {\bf Secure Expert Aggregation.}
For high accuracy, the expert aggregation scheme needs to assign different weights to different experts depending on their relevance. 
For IFC, the expert aggregation needs to be performed efficiently at run-time in a way that prevents the weights from leaking information from inaccessible domains.

\noindent {\bf Scalability and Modularity.} 
The traditional MoE architecture is only designed for a relatively small number of experts.
To allow scalability to a larger number of security domains, we need to carefully design the expert model architecture so that each expert has a relatively small number of parameters. For the same reason, instead of activating all accessible domains, we want our gating function to only activate a small number of the (top-$k$) most relevant experts. 
Finally, to efficiently support dynamic changes in security domains, gating decisions and aggregation methods should be able to quickly handle adding or removing security domains without re-training.

The security, modularity, and scalability constraints above significantly limit the information that can be used to make gating/aggregation decisions, the size of each expert, and the number of experts that can be activated ($k$). It is a significant challenge to achieve high accuracy under this constrained environment. In the following section, we describe our IFC design for LMs and show that our proposed design can significantly improve the accuracy while enforcing non-interference compared to a pre-trained model.



%% file: our-solution.tex
\vspace{-0.1in}
\section{IFC for Transformer Language Models}
\label{sec:ifc-nlp}
\vspace{-0.1in}

In this section, we describe how to adapt an existing Transformer architecture in language models (LMs) for information flow control (IFC). 
To ensure non-interference efficiently, we carefully choose an expert model architecture and develop secure gating and aggregation techniques for both cases when the domain of an input is known and unknown.

\vspace{-0.1in}
\subsection{Background: Transformer-based LM}
\vspace{-0.1in}

This paper focuses on language models which perform the next token prediction---the task of predicting the next most probable \emph{token} for any given sequence of tokens. For language modeling, the accuracy is measured with \emph{perplexity}, an exponentiation of the negative log likelihood of the text. Lower perplexity implies better performance.

\begin{figure}
    \centering
\includegraphics{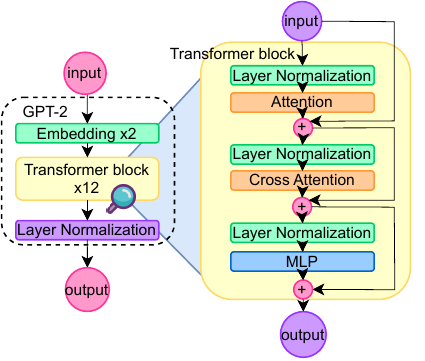}
\vspace{-0.10in}
    \caption{\small A typical Transformer-based language model (LM) architecture (based on GPT-2).}
    \label{fig:transformer}
\vspace{-0.2in}
\end{figure}

The current state-of-the-art language models use Transformers~\cite{vaswani2017attention, radford2018improving,devlin2018bert,yang2019xlnet,liu2019roberta,raffel2020exploring,zhang2022opt,black2022gpt}. 
Figure~\ref{fig:transformer} shows a typical decoder-only Transformer model architecture (GPT-2~\cite{radford2018improving}).
The model takes input texts represented as a sequence of tokens. These input tokens are passed through embedding layers, followed by Transformer blocks and a normalization layer. Each of the Transformer blocks consist of attention layers and an MLP layer, interspersed with normalization layers.

\begin{figure*} 
\begin{center}
\includegraphics{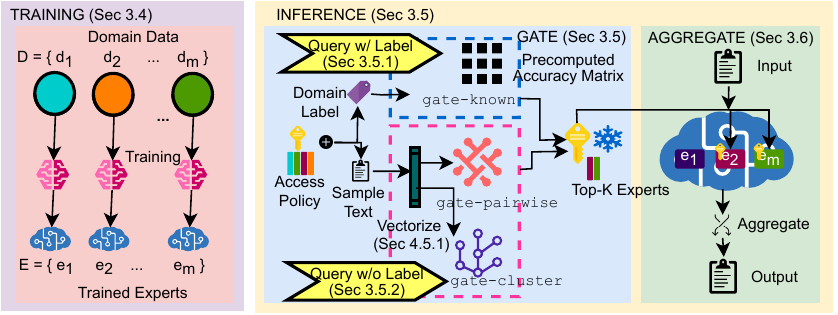}
\caption{\small The overview of the proposed Parametric IFC scheme for a Transformer-based language model.}
\label{fig:scheme}
\end{center}
\vspace{-0.3in}
\end{figure*}

\vspace{-0.1in}
\subsection{Overview of IFC for Parametric LM}
\label{ssec:overview}
\vspace{-0.05in}

Figure~\ref{fig:scheme} shows a high-level overview of our scheme.
During \textbf{domain-aware training}, we start with a baseline model pre-trained with public data. We fine-tune the pre-trained model with data from each of the $m$ security domains and get $m$ experts $e_1 ... e_m$. 
Training experts is done by fine-tuning only a subset of the parameters using data from each security domain.
These modified parameters of the model constitute the ``experts'', while the rest of the parameters remain unchanged. The effect of the data from each security domain is limited to parameters within one expert, thus enforcing parameter isolation (Section~\ref{sec:operationalization}). We discuss different expert architecture choices in Section~\ref{sec:exp-arch}.

Our \textbf{access-policy-aware inference} consists of two main steps: gating and aggregation. The {\bf gating} function selects the top-$k$ most relevant experts the user can access by taking in two (or three) inputs: a) the user $u_i$'s access policy $a_i$, b) a sequence of text tokens $\pmb{x} = \{{\pmb{x_1} ... \pmb{x_T}}\}$, and c) an \emph{optional} label $l$ that indicates which of the $m$ domains the input is closest to content-wise. The \textbf{aggregation} step uses the selected $k$ experts $\{e_1$ ... $e_k\}$ to produce a single high-quality output. We explore two approaches for aggregation: aggregating the expert outputs and aggregating the expert parameters.

Providing the domain label $l$ can be straightforward and useful in many cases. For example, when working on a particular project, other documents from the same project are likely to be the most relevant and may all be in the same security domain. In such cases, we use a gating algorithm called \texttt{gate-known}, which utilizes a pre-computed accuracy matrix to determine the top-$k$ experts accessible as per $a_i$ that are most relevant to domain $l$.
When the domain label is not provided, we determine the most relevant domains based on the previous input tokens (\texttt{gate-pairwise} and \texttt{gate-cluster}). These algorithms sample some tokens from the user's input (\emph{sample text}), convert it to a vector, and find the top-$k$ best matching experts to this sample vector through either a \emph{pairwise} distance comparison or a \emph{clustering} algorithm.

\vspace{-0.1in}
\subsection{Expert Model Architecture Choices}
\label{sec:exp-arch}
\vspace{-0.1in}

There are many potential choices for experts. For resource efficiency, it is better to only fine-tune a small number of parameters. For accuracy, it may be better to train more parameters. Below are the different options we explored:

\begin{enumerate}[noitemsep, leftmargin=*, topsep=0pt]
    \item \textbf{Fully Fine-tune.} For the best accuracy, we can fine-tune all parameters of separate copies of the model using data from each security domain to create experts.
    
    \item \textbf{MLP.} We freeze all layers except for the MLP layers in each Transformer (Encoder/Decoder) layer. Each domain expert only differs in its MLP weights.

    \item \textbf{First $\frac{N}{2}$ Decoder Layers.} We fine-tune only the first $\frac{N}{2}$ of the $N$ Decoder blocks and freeze the rest.
    
    \item \textbf{Last $\frac{N}{2}$ Decoder Layers.} We freeze the first half of the Decoder blocks and fine-tune the rest.
    
    \item \textbf{Adapter.} 
    Adapters are blocks consisting of a few layers that are added to a pre-trained Transformer for efficient model fine-tuning~\cite{houlsby2019parameter}. 
    We insert additional adapter blocks and only fine-tune them with the security domain data. 
\end{enumerate}

We evaluated these expert architecture options in Section~\ref{sec:evaluation} using GPT-2 as our representative model, and chose the adapter as the main option in this study. For this setup, we place an adapter block right before the MLP layer in each of the Transformer blocks and only train the adapters for each domain. While the expert architecture is important for storage overhead and fine-tuning time, the rest of the scheme can work for any expert architecture choice.

\vspace{-0.1in}
\subsection{Domain-Aware Training}
\vspace{-0.05in}

During training, we take the text data in each of the $m$ security domains in $D$, and train one expert per security domain (Algorithm~\ref{alg:train}). Note that the training algorithm that produces an expert $e_j$ only depends on the domain data for that expert, i.e., $d_j$. This means that the experts are trained independent of each other, and can be added or removed from the final model as needed, allowing us to make our model \emph{modular}. 

\SetNlSty{textbf}{(}{)}
\begin{algorithm}[b]
\small
\SetKwFunction{KwTrain}{train}
\SetKwFunction{KwEnsemble}{aggregate}
\KwIn{Training Data $D = \{d_1, d_2$ ... $d_m\}$}
\KwOut{Trained Experts $e_1$ ... $e_m$}
\For{$j \gets 1$; $j \leq m$; $j\mathrel{+}= 1$}{
    $e_j \gets$ \KwTrain{$d_j$}
}
\caption{\small Training algorithm for IFC.}
\label{alg:train}
\end{algorithm}

\vspace{-0.1in}
\subsection{Access-Policy-Aware Gating}
\label{sec:gate-k}
\vspace{-0.05in}

In order to optimize accuracy and scalability, we only activate a small number ($k$) of experts that are accessible as per the user's access policy $a_i$, and relevant to the given input $\pmb{x}$. 
If we use \emph{all} accessible experts, the outputs from irrelevant experts can hurt the accuracy. Also, a large number of activated experts can incur significant resource overhead. Thus, we limit the aggregation to the top-$k$ closest experts.

\subsubsection{Known Domain Label}
\vspace{-0.05in}
Algorithm~\ref{alg:known} (\texttt{gate-known}) shows how we choose the top-$k$ experts when the domain label $l$ is given by the user.
In the offline phase of the algorithm, we evaluate each of the domain experts ($e_1$ ... $e_m$) on a small amount of held out text from each security domain ($w_1$ ... $w_m$) to obtain an all-to-all accuracy matrix ($M$) (Line 1). The rows of $M$ pertain to the security domain of the held out text, and the columns pertain to the experts that is used to evaluate the text. This allows us to see how well each expert performs on text from each domain.

During inference, when the domain label $l$ is provided with the user's access policy $a_i$, \texttt{gate-known} first ensures that $l$ is one of the user's accessible domains\footnote{To ensure that a user cannot learn any information about an inaccessible domain, an error is returned and the inference is aborted if a user provides the label $l$ for an inaccessible domain.} (Lines 2-3) and uses $l$ to select the most relevant domains from the accuracy matrix $M$. The function then uses the access policy $a_i$ as a mask to filter out the inaccessible experts. This step is important to ensure inaccessible experts are removed from the top-$k$ consideration. The final outcome is a list of accessible experts ranked by their performance on domain $l$ (Line 4). We then select the top-$k$ experts from this ordered list. 

Note that this gating algorithm only uses information from accessible domains, and the gating decisions do not leak information about inaccessible domains.

\begin{algorithm}[t]
\small
\SetKw{KwHy}{Hyperparameters:}
\SetKwFunction{KwPerp}{perplexity\_matrix}
\SetKwFunction{KwAbort}{abort}
\SetKwFunction{KwAccessible}{accessible}
\SetKwFunction{KwNot}{not}
\SetKwFunction{KwGate}{gate}
\SetKwFunction{KwTopK}{top\_k}
\SetKwFunction{KwEnsemble}{aggregate}
\KwIn{User input $\pmb{x_1}$ ... $\pmb{x_T}$, access policy $a_i$, domain label $l \in D$}
\KwOut{Model output $o_c$ ... $o_T$}
\nonl \KwHy{$k$} \\
\tcp{Offline Phase}
$M$ $\gets$ \KwPerp{$\{e_1 ... e_k\}$, $\{w_1 ... w_k\}$} \\
\tcp{Online Phase}
\If {\KwNot \KwAccessible{l}}{\KwAbort;}
$e_1$ ... $e_k$ $\gets$ \KwTopK{\KwGate{$a_i$, $l$, $M$}} \\
\For{$t \gets 1$; $t \leq T$; $t\mathrel{+}= 1$}{
    $o_t$ $\gets$ \KwEnsemble{$\pmb{x_t}$, $e_1$ ... $e_k$}
}
\caption{\small Gating w/ a domain label (\texttt{gate-known}).} 
\label{alg:known}
\end{algorithm}


\vspace{-0.1in}
\subsubsection{Unknown Domain Label}
\vspace{-0.05in}
When the domain label $l$ is not provided, we use a part of the user's input (specifically, the first $c$ tokens) as a sample text to estimate top-$k$ experts that are the most relevant. 
The simplest way to determine relevant experts is to run the sample text through each expert and compare the resulting perplexity. However, this approach is too expensive for a real-time gating function.
Instead, we compare the vectorized sample text with the vectorized representation of each domain (Algorithm~\ref{alg:unknown}). First, we convert the sample text $\{\pmb{x_t}$ ... $\pmb{x_{t+c-1}}\}$ to a vector representation (Line $2$). This vector is passed to the \texttt{gate} function, which outputs the top-$k$ experts from the accessible domains $a_i$. We explore two alternatives of the \texttt{gate} function in Line $3$, provided by \texttt{gate-pairwise} (Algorithm~\ref{alg:pairwise}) and \texttt{gate-cluster} (Algorithm~\ref{alg:cluster}).
As we will elaborate later, both gating functions are carefully designed to ensure non-interference.  
Once these experts $\{e_1$ ... $e_k\}$ are determined, the tokens after the sample text, i.e., $\pmb{x_{>t+c}}$, and the selected experts $\{e_1$ ... $e_k\}$ are fed into an \texttt{aggregate} function (Section~\ref{sec:ensemble}) to produce an output $o_{>t+c}$ (Lines $4$-$5$). The most relevant $k$ experts are re-evaluated after every $r$ tokens in the user's input, in order to account for the domain shift in the user's inputs over time. 


\begin{algorithm}[t]
\small
\SetKw{KwHy}{Hyperparameters:}
\SetKwFunction{KwGate}{gate}
\SetKwFunction{KwVector}{vectorize}
\SetKwFunction{KwEnsemble}{aggregate}
\KwIn{User input $\pmb{x_0}$ ... $\pmb{x_T}$, access policy $a_i$}
\KwOut{Model output $o_c$ ... $o_T$}
\nonl \KwHy{$r$, $c$, $k$} \\
\For{$t \gets 1$; $t \leq T$; $t\mathrel{+}= r$}{
    $\pmb{v^{\prime}}$ $\gets$ \KwVector{$\pmb{x_t}$ ... $\pmb{x_{t+c-1}}$} \\
    $e_1$ ... $e_k$ $\gets$ \KwGate{$\pmb{v^{\prime}}$, $a_i$} \\
    \For{$t^{\prime} \gets t+c$; $t^{\prime} \leq t$; $t^{\prime} \mathrel{+}= 1$}{
        $o_{t^{\prime}}$ $\gets$ \KwEnsemble{$\pmb{x_{t^{\prime}}}$, $e_1$ ... $e_k$}
   }
}
\caption{\small Gating w/o domain labels (\texttt{gate-unknown}).}
\label{alg:unknown}\end{algorithm}


The \texttt{vectorize} function (Line 2, Algorithm~\ref{alg:unknown}) converts a text to its vector representation. 
Popular methods to vectorize text include simpler approaches such as Bag of Words~\cite{bagofwords} (and its variants like TF-IDF and BM25), n-gram vectors~\cite{ngram}, and more sophisticated approaches involve encoder models like BERT~\cite{devlin2018bert}. While easier and more efficient to implement, the simpler techniques have many shortcomings~\cite{embeddingShortcomings} (e.g., Word2Vec cannot work with tokens outside of the training vocabulary). 
Thus, we use pre-trained BERT~\cite{devlin2018bert}, which is an encoder model that converts each text-token to a $768$-dimensional embedding vector. For multiple tokens, we compute an average over all tokens in the text to obtain a single $768$-dimensional vector. For the user's sample text, BERT is run at the inference time to obtain its embedding vector. For each security domain, we pre-compute a representative embedding vector offline with BERT using its training data. 
The \texttt{gate-pairwise} function (Algorithm~\ref{alg:pairwise}) picks the top-$k$ most relevant experts by comparing the pairwise distance between the vectorized sample text and the representative vector of each accessible domain. 
First, the representative vector $\pmb{v} = \{\pmb{v_1}, \pmb{v_2}$ ... $\pmb{v_m}\}$ for each domain is calculated offline (Line 1-2). 
During inference, a pairwise cosine similarity score ($\varphi$) between the sample text vector $\pmb{v^{\prime}}$ and every single domain vector that is accessible to the user $\bigcup_{n \in a_i}\{\pmb{v_n}\}$ is calculated to get a list of similarity scores $\bigcup_{n \in a_i}\{\varphi_n\}$ (Line 4).
Note that the similarity scores are not computed with inaccessible domains to prevent information leakage. 
A higher similarity between $\pmb{v^{\prime}}$ and any domain vector $\pmb{v_n}$ suggests that the user's text is similar to that particular domain, and the corresponding expert is likely to be useful. We then adjust the resulting cosine similarity rankings of the accessible domains using the \texttt{score} function (discussed below) to get a list of adjusted similarity scores ($\bigcup_{n \in a_i}\{\varphi^{\prime}_n\}$). We select experts with the top-$k$ highest adjusted similarity scores (Line 6).
\begin{algorithm}[t]
\small
\SetKw{KwHy}{Hyperparameters:}
\SetKwFunction{KwVector}{vectorize}
\SetKwFunction{KwPairwise}{pairwise}
\SetKwFunction{KwGate}{gate}
\SetKwFunction{KwScore}{score}
\SetKwFunction{KwTopK}{top\_k}

\KwIn{$c$ tokens from user input {$\pmb{x_t}$ ... $\pmb{x_{t+c-1}}$}, access policy $a_i$}
\KwOut{Top-$k$ experts $e_1$ ... $e_k$}
\nonl \KwHy{$k$, $c$} \\
\tcp{Offline Computation}
\For{$j \gets 0$; $j \leq m$; $j\mathrel{+}= 1$}{
    $\pmb{v_j}$ $\gets$ \KwVector{$d_j$}
} 
\tcp{Online Phase}
\textbf{def} \KwGate{$\pmb{v^{\prime}}$, $a_i$} \\
\Indp 
$\bigcup_{n \in a_i}\{\varphi_n\}$ $\gets$ \KwPairwise{$\pmb{v^{\prime}}$, $\bigcup_{n \in a_i} \{\pmb{v_n}\}$} \\
$\bigcup_{n \in a_i}\{\varphi^{\prime}_n\}$ $\gets$ \KwScore{$\bigcup_{n \in a_i}\{\varphi_n\}$} \\
$e_1$ ... $e_k$ $\gets$ \KwTopK{$\bigcup_{n \in a_i}\{\varphi^{\prime}_n\}$}
\caption{\small Pairwise gating function (\texttt{gate-pairwise}).}
\label{alg:pairwise}
\end{algorithm}

\begin{algorithm}[t]
\small
\SetKw{KwHy}{Hyperparameters:}
\SetKwFunction{KwVector}{vectorize}
\SetKwFunction{KwClusterCompute}{compute\_clusters}
\SetKwFunction{KwClusterCenters}{cluster\_centers}
\SetKwFunction{KwClusterNearest}{nearest\_cluster}
\SetKwFunction{KwTopK}{top\_k}
\SetKwFunction{KwGate}{gate}
\SetKwFunction{KwScore}{score}
\SetKwFunction{KwPairwise}{pairwise}
\KwIn{$c$ tokens from user input {$\pmb{x_t}$ ... $\pmb{x_{t+c-1}}$}, access policy $a_i$}
\KwOut{Top-$k$ experts $e_1$ ... $e_k$}
\nonl \KwHy{$k$, $c$, $s$} \\

\tcp{Offline Computation}
\For{$j \gets 0$; $j \leq m$; $j\mathrel{+}= 1$}{
    $\pmb{v_j}$ $\gets$ \KwVector{$d_j$}
}
$C_0$ ... $C_s$ $\gets$ \KwClusterCompute{$\pmb{v_1}$ ... $\pmb{v_m}$} \\

\tcp{Online Phase}
\textbf{def} \KwGate{$\pmb{v^{\prime}}$, $a_i$} \\
\Indp 
$c_1$ ... $c_s$ $\gets$\KwClusterCenters{$C_1$ ... $C_s$, $a_i$} \\
$C_i$ $\gets$ \KwClusterNearest{$\pmb{v^{\prime}}$, $c_1$ ... $c_s$, $a_i$} \\
$\bigcup_{n \in C_i \cap a_i}\{\varphi_n\}$ $\gets$ \KwPairwise{$\pmb{v^{\prime}}$, $\bigcup_{n \in C_i \cap a_i}\{\pmb{v_n}\}$} \\
$\bigcup_{n \in C_i \cap a_i}\{\varphi^{\prime}_n\}$ $\gets$ \KwScore{$\bigcup_{n \in C_i \cap a_i}\{\varphi_n\}$} \\
$e_1$ ... $e_k$ $\gets$ \KwTopK{$\bigcup_{n \in C_i \cap a_i}\{\varphi^{\prime}_n\}$}
\caption{\small Cluster gating function (\texttt{gate-cluster}).}
\label{alg:cluster}
\vspace{-0.02in}
\end{algorithm}


%
The \texttt{gate-cluster} function (Algorithm~\ref{alg:cluster}) is a more efficient alternative than \texttt{gate-pairwise} which performs a linear search over $m$ domains and scales poorly with a large $m$. Instead,
\texttt{gate-cluster} introduces a hierarchical search, being more suitable when $m$ is large.
During the offline phase, \texttt{gate-cluster} clusters the domain vectors into $s$ clusters of similar domains $D = \{C_0$ ... $C_s\}$, where $\bigcap_{j = 0}^{s} C_j = \phi$ and $\bigcup_{j = 0}^{s} C_j = D$ (Line 3). During inference, the \texttt{gate} function first computes the centers $c_1$ ... $c_s$ of each cluster based on the accessible domains (Line 5). These cluster centers are then used to calculate the cosine similarity with the sample text vector $\pmb{v^{\prime}}$ to find the closest cluster with at least one accessible domain (Line 6). While the clusters themselves are allocated offline, the cluster centers for each of the clusters need to be computed online to ensure that inaccessible domains do not influence the closest cluster selection. Within this closest cluster $C_i$, we compute the pairwise cosine similarity between $\pmb{v^{\prime}}$ and each accessible domain vector in the cluster $C_i$ to get a list of scores $\bigcup_{n \in C_i \cap a_i}\{\varphi_n\}$ (Line 7). We again adjust the scores using the \texttt{score} function and pick $k$ experts with the highest scores (Line 9). 

The \texttt{score} function (Algorithm~\ref{alg:pairwise}, Line 5; Algorithm~\ref{alg:cluster}, Line 7) adjusts the similarity scores by considering the quality of each expert.
The initial cosine similarity score only captures the semantic similarity between the sample text and the domain's text, but fails to take into account how well the expert corresponding to the domain was trained. Even when the cosine similarity is low (the domain is not close to the sample text), experts pertaining to domains with a large number of samples sometimes provide a better output as they are trained better. 
The \texttt{score} function adjusts the cosine similarity score $\varphi_n$ by taking into account the amount of training data for each domain (Equation~\ref{eq:similarity}).
We introduce a term $S_n \in [0,1]$, which represents the fraction of training samples out of all accessible samples that belong to domain $n$. $S_n$ is close to 1 for domains with more training data.
We add $S_n$ multiplied with  $\lambda \in [0,1]$ to get the adjusted score, where $\lambda$ is a hyper-parameter that controls the influence of the domains' training data size. If $\lambda$ is close to 1, domains with more samples are preferred.
\begin{align}
S_n = \frac{\text{count}(d_n)}{\sum_{j = 1}^{len(a_i)}\text{count}(d_j)}, \quad
\varphi^{\prime}_n = \text{score}(\varphi_n) = \varphi_n + \lambda S_{n}
\label{eq:similarity}
\end{align}

\vspace{-0.1in}
\subsection{Expert Aggregation} 
\label{sec:ensemble}
\vspace{-0.05in}

When the top-$k$ experts are selected, they are aggregated to produce a single output (Algorithm~\ref{alg:known}, Line 6; Algorithm~\ref{alg:unknown}, Line 5). We explore two ways of aggregation: ensembling the expert outputs and merging the expert parameters.

The first option is to \textbf{\emph{ensemble the outputs}} from the $k$ experts. We first run the user's input at time $t$ ($\pmb{x_t}$) through the selected $k$ experts ($e_1$ ... $e_k$), and combine the resulting $k$ outputs ($o_{t,1}$ ... $o_{t,k}$) to give a final output $o_t$ at time $t$. We employ an approach similar to DeMix~\cite{gururangan2021demix}, which exploits Bayesian probabilities to estimate the likelihood of each of the $k$ experts being a good fit for the input $\pmb{x_t}$ (Equation~\ref{eq:ensemble1}) and uses these likelihoods as weights to aggregate the $k$ outputs (Equation~\ref{eq:ensemble2}). 
The likelihood, expressed as a posterior probability over domains, can be calculated using Bayes' rule~\cite{gururangan2021demix}:
\begin{align}
\label{eq:ensemble1}
    P(d_t = j | \pmb{x_t}) &= \frac{P(\pmb{x_{<t}} | d_t = j) \times P(d_t = j)}{P(\pmb{x_{<t}})} \\
    &= \frac{P(\pmb{x_{<t}} | d_t = j) \times P(d_t = j)}{\sum_{j^{\prime} = 1}^{k}P(\pmb{x_{<t}} | d_t = j^{\prime}) \times P(d_t = j^{\prime})}.
\end{align}
We assume a uniform domain prior (i.e., uniform $P(d_t = j)$) --- that is, no one domain is inherently more likely than another. Our experts already output the probability of tokens conditioned on the respective experts ($P(\pmb{x_{<t}} | d_t = j)$). These domain likelihoods are used to aggregate the outputs of each of the $k$ experts to get the final output:
\begin{equation}
    o_t = \sum_{j = 1}^{k} o_{t,j} \times P(d_t = j | \pmb{x_t}).
\label{eq:ensemble2}
\end{equation}
This method does not require training and uses information only from accessible domains, allowing easy addition/removal of domains while ensuring non-interference.   

Another option is to \textbf{\emph{merge the parameters}} of the $k$ experts and feed the user's input to this combined expert. This has the advantage of reducing the number of experts that need to be run at inference. The weights that we use for parameter merging are simply a normalized version of the top-$k$ scores returned by our gating algorithms. This parameter aggregation approach is inspired by the recent studies on combining parameters in model training \cite{wortsman2022model,li2022branch, izmailov2018averaging}.
While the approach is similar, we apply the parameter merging at inference time and show that it can also be used to combine multiple experts.
With regards to IFC, since the top-$k$ experts are all accessible according to the user's access policy, merging parameters from these already accessible experts does not lead to any leakage from inaccessible security domains.

%% file: experiments.tex
\vspace{-0.1in}
\section{Evaluation}
\label{sec:evaluation}
\vspace{-0.1in}

\subsection{Experimental Setup}
\vspace{-0.05in}
\noindent{\bf Baseline NLP Models.}
We use GPT-2~\cite{radford2018improving} and OPT~\cite{zhang2022opt} as two representative Transformer-based language models to evaluate our approach. 
Given the architecture-agnostic properties of our approach, we believe that the proposed information flow control framework can also be applied to other state-of-the-art Transformer-based models~\cite{devlin2018bert,yang2019xlnet,liu2019roberta,raffel2020exploring}.

\noindent{\bf Datasets.}
We use two datasets, the Pushshift.io dataset~\cite{baumgartner2020pushshift}, and Codeparrot's Github-Code dataset~\cite{codeparrot}. The former is used to train GPT-$2$, while the latter is used for OPT. Pushshift~\cite{baumgartner2020pushshift} is a pre-existing, publicly available third-party dataset composed of an actively maintained collection of all Reddit comments and posts from $2005$. We simulate security domains by treating different subreddits as different domains. While we only use public content from Pushshift.io, a similar platform may want to have access control for different groups, so that each user has access to only a subset of groups. The setup is also representative of a general case where the security domain and the content have a correlation (e.g., different projects forming their own security domain). We set the number of domains to $m = 50$ for our experiments with Pushshift, which corresponds to $50$ different subreddits. 
For the Pushshift experiments, we use all comments from the $50$ subreddits over the time-span of August $2021$ to August $2022$. The domain sizes range from $\approx19$k tokens in the smallest domain to $\approx500$M tokens in the largest. 
%

Our second dataset is Codeparrot's Github-Code dataset~\cite{codeparrot}. This dataset is a collection of open-source code from different repositories spanning various licenses on GitHub. We use data from MIT licensed repositories from $5$ different programming languages: Python, JavaScript, C, C++ and Java. We take the largest MIT licensed repositories from these languages to get a total of $79$ domains, where each security domain spans a single code repository. More details of the datasets are in Appendix~\ref{sec:appendix-dataset}.   



%

\noindent{\bf Baseline Schemes.}
We use two baselines to compare our algorithms with: a) pre-trained GPT-2/OPT and b) GPT-2/OPT fine-tuned on the data from all domains combined together. The first ensures non-interference, as none of the inaccessible domains' data is used for training and thus cannot be leaked during inference. 
The second represents an insecure baseline, which provides the best case accuracy that can be achieved using all access-controlled data for fine-tuning.

Figure~\ref{fig:baseline} shows the perplexity when the test sets from each of the domains are evaluated for the baselines a) and b) for the Pushshift dataset. We see that the pre-trained model does not perform well, showing the average perplexity over all domains of $55.5$. The average perplexity is significantly improved to $29.8$ for the model fine-tuned on all $50$ domains. This suggests that using in-domain data when possible can significantly improve model accuracy. For all experiments that follow, improvement is shown as the normalized perplexity over the pre-trained model. 

\begin{table}[]
\footnotesize
\begin{tabular}{@{}|c|c|@{}}
\toprule
Param     & Description                                                            \\ \midrule
$m$       & Number of security domains                                             \\
$k$       & Number of experts to activate \\
$r$       & \makecell{Number of tokens to process \\ after which we re-evaluate top-$k$ experts} \\
$c$       & Number of tokens in sample text                             \\
$s$       & Number of clusters used for \texttt{gate-cluster}     \\
$\lambda$ & Factor by which we penalize the experts of small domains             \\ \bottomrule
\end{tabular}
\caption{\small Description of the hyper-parameters.}
\label{tab:hyperparam}
\end{table}

\noindent{\bf Parametric IFC schemes and hyper-parameters.}
We evaluate the three proposed algorithms for gating: \verb|gate-known|, \verb|gate-pairwise|, and \verb|gate-cluster| together with output ensembling as our aggregation scheme (unless noted otherwise).
For \verb|gate-known|, we use the subreddit group/code-repository that the query was originally from as the user-provided domain label $l$. 
Table~\ref{tab:hyperparam} summarizes the hyper-parameters for our IFC schemes. 
We use the following as the default values: $k = 3$, $r = 500$k, $c = 10$k, $s = 10$, and $\lambda = 0.4$. An ablation study over these hyper-parameters is provided in Appendix~\ref{sec:appendix-ablation}

\begin{figure}
    \centering
\includegraphics{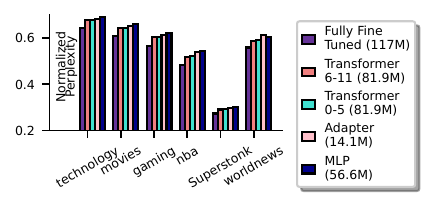}
    \vspace{-0.2in}
    \caption{\small Accuracy of different expert model architectures (Section~\ref{sec:exp-arch}). The number of trainable parameters is shown in the legend. The y-axis shows the perplexity normalized with respect to pre-trained only model. A lower value represents a bigger drop in perplexity, and thus a larger improvement.}
    \label{fig:exp-arch}
    \vspace{-0.1in}
\end{figure}

\vspace{-0.1in}
\subsection{Impact of Expert Architecture}
\vspace{-0.05in}

To evaluate different expert architecture choices (discussed in Section~\ref{sec:exp-arch}), we trained all $5$ architectures with GPT-$2$ on the $5$ largest domains in the Pushshift dataset, and evaluated the performance of the respective experts on the validation sets of said domains. Figure~\ref{fig:exp-arch} shows the results. The y-axis represents the normalized perplexity with respect to the pretrain-only model. A lower value represents a higher drop in perplexity and therefore a larger improvement in accuracy. 

We notice that all five architecture options lead to a $30\%$ to $70\%$ perplexity improvement depending on the domain, with the fully fine-tuned expert outperforming other architectures, albeit by a maximum margin of $0.1\%$. However, as seen in the legend, the fully fine-tuned expert also uses considerably more trainable parameters than the other architectures. In contrast, the adapter uses the smallest number of trainable parameters, but still performs quite well (only $0.1\%$ worse in the worst case scenario). Given this, for the following experiments, we use adapters as our expert architecture. 

\begin{figure}
\centering
\includegraphics{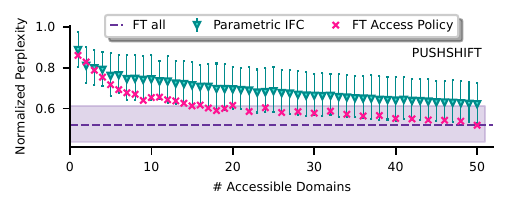}
\vspace{-0.3in}
  \caption{\small The normalized perplexity as a function of the number of accessible domains for our parametric IFC scheme. Perplexity is normalized to the pre-trained model, and each point and error bar represents the geometric mean/standard deviation over all domains $\times$ gating algorithms $\times$ access policies. ``FT Access Policy'' represents perplexities evaluated for a random subset of access policies where the model was fully fined tuned on all \emph{accessible data} for each policy. The horizontal line and shaded regions represent the geometric mean/standard deviation of the insecure, fully fine-tuned model.} 
  \label{fig:all}
\vspace{-0.2in}
\end{figure}

\begin{figure*}
\centering
\includegraphics{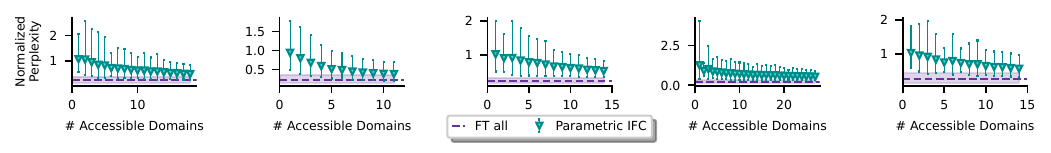}
\vspace{-0.3in}
  \caption{\small The normalized perplexity as a function of the number of accessible domains for each programming language in Codeparrot.}
  \label{fig:github-all}
\vspace{-0.2in}
\end{figure*}

\vspace{-0.1in}
\subsection{Model Accuracy}
\vspace{-0.05in}

To understand the accuracy impact of the IFC scheme, we evaluate the perplexity of the language model using the test sets of all $50$ domains in the Pushshift dataset and $79$ domains in the Codeparrot dataset under a large number of access policies. For this study, we randomly selected $500$ access policies from all $2^{50}$ possible combinations for the Pushshift dataset, and $790$ out of $2^{79}$ for the Codeparrot dataset. The policies cover all possible number of accessible domains ($1$ to $50$ and $1$ to $79$ respectively). There are about $10$ policies for a given number of accessible domains. Since the $79$ domains in Codeparrot span $5$ different programming languages, we depict our results for each programming language individually. 

\begin{figure}
\centering
  \includegraphics{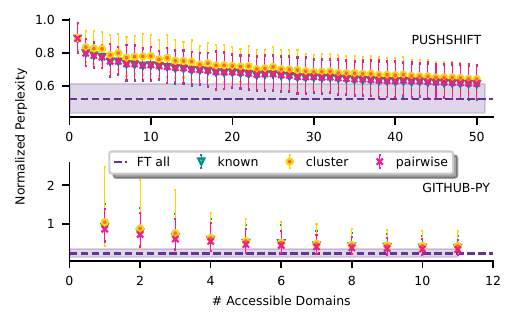}
\vspace{-0.3in}
  \caption{\small Impact of different gating algorithms.}
  \label{fig:gating}
\vspace{-0.2in}
\end{figure}
\begin{figure}
\centering
  \includegraphics{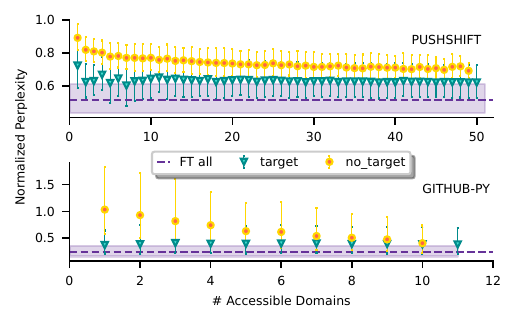}
\vspace{-0.3in}
  \caption{\small Effect of including vs. excluding target domain.}
  \label{fig:target}
\vspace{-0.2in}
\end{figure}
\begin{figure}
\centering
  \includegraphics{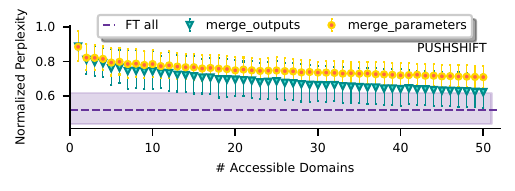}
  \caption{\small The accuracy comparison between the two expert aggregation methods: output ensembling and parameter merging (GPT-2 for the Pushshift dataset).}
  \label{fig:agg}
\vspace{-0.25in}
\end{figure}

Each of the graphs in Figures~\ref{fig:all},~\ref{fig:github-all},~\ref{fig:gating},~\ref{fig:target}, and~\ref{fig:agg} shows how the normalized perplexity varies as more domains become accessible for the Pushshift dataset and repositories in the CodeParrot dataset. The y-axis shows the perplexity normalized to the pre-train only model (lower is better). 
We divide all the evaluated access policies into bins indicating the number of accessible domains. 
Each point in the figures shows the geometric mean over multiple access policies and multiple domains for a given number of accessible domains.  We also show the geometric mean of the normalized perplexity of the fully fine-tuned (insecure) model, which uses the training data from all domains. The shaded regions and error bars represent one geometric standard deviation.



\noindent\textbf{Impact of the Number of Accessible Domains.}  
Figures~\ref{fig:all} and~\ref{fig:github-all} show that over all possible gating schemes and access policies, the accuracy with IFC improves as the number of accessible domains increases. Even with just one accessible domain in the Pushshift dataset (Figure~\ref{fig:all}), the perplexity is reduced by $11.5\%$ on average compared to the model that is pre-trained on public data only. When all domains are accessible, the perplexity is reduced by about $38\%$, close to the ideal $48\%$ improvement of the fully fine-tuned model. For the Codeparrot dataset (Figure~\ref{fig:github-all}), we see an improvement of $44-62\%$ when all domains are accessible, with the fully fine-tuned model at roughly $75\%$ improvement for all coding languages. Note again that full fine-tuning is an insecure baseline that leaks data from inaccessible domains.

Figures~\ref{fig:all} and~\ref{fig:github-all} show higher perplexities when only few domains are accessible. The low accuracy is because the accessible experts are less likely to be useful for a given input when there are only a few, not because our approach is ineffective for a small number of domains. To illustrate this, we fine-tuned a separate model for each access policy (for $50$ out of the $500$ policies in Figure~\ref{fig:all}). We see the same trend of lower accuracy for fewer accessible domains, with accuracy improving as more domains become accessible.



\noindent\textbf{Impact of Gating Algorithm.} 
Figure~\ref{fig:gating} separates the results from Figure~\ref{fig:all} by the gating algorithm used. This result shows that all gating schemes are effective at selecting relevant experts.
\verb|gate-known|, which has extra knowledge of the domain label, performs the best as expected.
Also, \verb|gate-pairwise| performs better than \verb|gate-cluster|;
however, the differences are relatively small. The average difference is $3\%$ and $9\%$ between the two for both the Pushshift dataset and the Python subset of the Codeparrot dataset respectively. 

%



\noindent\textbf{Including vs. Excluding Target Domain.} 
We additionally evaluate settings when the security domain that is the most relevant to the user input (i.e., target domain) is \emph{inaccessible}.
We do so by excluding the subreddit/repository the input query was selected from.
This simulates the case where the user queries a new domain that the model is not trained on yet.
Figure~\ref{fig:target} demonstrates the differences in the two cases where the target domain is accessible or inaccessible. 

When the number of accessible domains are small, the gap between the two cases (target domain included vs. excluded) is large.
This is because when the target domain is excluded, the model cannot find an expert that directly matches the user query's domain.
However, the gap decreases as more domains become accessible, showing the proposed scheme's ability to aggregate less related experts to improve the output when the most relevant domain is unavailable.
Interestingly, the maximum improvement ($\approx 40$\% and $\approx 63\%$ for Pushshift and Python-Codeparrot, respectively) is only slightly lower (by $8\%$ and $4\%$ for Pushshift and Python-Codeparrot) when the target domain is excluded compared to when the target domain could be accessible. This shows that when enough experts are accessible, many non-target experts can almost fully compensate for the lack of the target expert. 

\noindent\textbf{Output vs. Parameter Aggregation.}
The results so far show that output ensembling works well as the expert aggregation scheme. 
In order to understand the potential to avoid running multiple experts, we also tested the parameter merging approach for GPT-2 with the Pushshift dataset.
Figure~\ref{fig:agg} compares the normalized perplexity of output ensembling and parameter merging. Here we see that the maximum improvement from output ensembling is $38\%$, whereas that from parameter merging is only $29\%$. Thus, while parameter merging can also improve perplexity over the pre-train only baseline and may be used as an efficient aggregation option, the result suggests that output ensembling is more effective, especially when a large number of domains are accessible.
We leave in-depth studies on parameter merging to future work.

\vspace{-0.1in}
\subsection{Performance and Resource Usage}
\vspace{-0.1in}

\textbf{Use Case Scenarios.}
The run-time of our scheme depends on the use case. 
Generative use cases can be roughly divided into two categories: single token generation (e.g., next word prediction) and batch token generation (e.g., document paraphrasing, question/answering, summarization, etc.). 

\begin{figure}
\begin{center}
\includegraphics{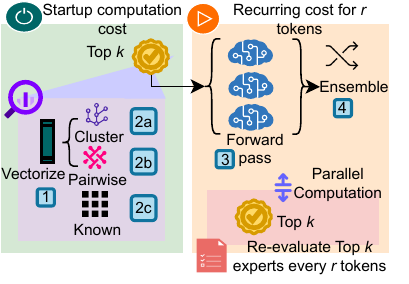}
\vspace{-0.1in}
\caption{\small A breakdown of the inference run-time when evaluating a sequence of input tokens. The startup part runs only once, while the recurring part runs for every sequence of $r$ tokens.} 
\label{fig:runtime}
\end{center}
\vspace{-0.31in}
\end{figure}

\noindent\textbf{Run-time Overhead.}
We show the flow of both use cases in Figure~\ref{fig:runtime} (here, we only evaluate output aggregation). 
In both cases, the run-time cost can be divided into two parts --- a startup cost and a recurring cost. The startup cost involves picking the initial top-$k$ experts for the first batch of the user provided tokens. The startup cost has to be incurred before we can run our model to provide predictions. The recurring cost involves evaluating input text tokens with the top-$k$ experts and aggregating the outputs. This process is repeated as long as the user keeps providing input. As discussed in Section~\ref{sec:ifc-nlp}, we re-evaluate the top-$k$ experts after every $r$ tokens to handle the changing nature of text . The re-evaluation can be done in parallel with the model execution for the current sequence of tokens and does not add additional latency. 



\begin{table*}
\scriptsize
\centering
\begin{tabular}{@{}ccccccc@{}}
\toprule
Component / Complexity & Variable & \multicolumn{5}{c}{Runtime (s) / Memory (MiB)} \\ \midrule
\multirow{4}{*}{\makecell{\includegraphics[scale=0.5,trim={0cm 0.11cm 0cm 0cm},clip]{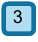} Forward Pass \\ O($n$)}} & \multirow{2}{*}{\makecell{Model Size \\ (1k Tokens)}} & Small & Medium & Large & XL & - \\
 &  & 7.45 / 612 & 13.8 / 1635 & 21.1 / 3175 & 27.5 / 6272 & - \\ \cmidrule(l){3-7}
 & \multirow{2}{*}{\makecell{Num Tokens \\ (GPT-$2$ Large)}} & 1 & 10 & 100 & 500 & 1k \\
 &  & 0.166 / 3618 & 0.354 / 3627 & 2.19 / 3251 & 10.4 / 3213 & 21.1 / 3175 \\ \cmidrule(l){2-7}

\multirow{2}{*}{\makecell{\includegraphics[scale=0.5,trim={0cm 0.11cm 0cm 0cm},clip]{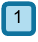} Vectorize \\ O($n$)}} & \multirow{2}{*}{Context Size} & \textless{}512 & 1024 & 2048 & 5120 & 10240 \\
 &  & 7.87E-03 / 1206 & 1.57E-02 / 2412 & 3.13E-02 / 2412 & 7.82E-02 / 2412 & 1.56E-01 / 2412 \\ \cmidrule(l){2-7}
 
\multirow{4}{*}{\makecell{\includegraphics[scale=0.5,trim={0cm 0.11cm 0cm 0cm},clip]{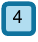} Output Aggregation \\O($n(2k+1)$)}} & \multirow{2}{*}{\makecell{Num Tokens \\ (3 Experts)}} & 1 & 5 & 512 & 1024 & - \\
 &  & 2.37E-04 / 260 & 3.13E-04 / 263 & 6.58E-03 / 554 & 1.31E-02 / 849 & - \\ \cmidrule(l){3-7} 
 & \multirow{2}{*}{\makecell{Num Experts \\ (1024 Tokens)}} & 2 & 3 & 4 & 5 & - \\
 &  & 9.73E-03 / 651 & 1.31E-02 / 849 & 1.64E-02 / 1045 & 1.97E-02 / 1241 & - \\
\cmidrule(l){2-7} 
 
\multirow{2}{*}{\makecell{\includegraphics[scale=0.5,trim={0cm 0.11cm 0cm 0cm},clip]{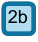} \texttt{gate-pairwise} \\ O($m*\log(m)$)}} & \multirow{2}{*}{Num Domains} & 100 & 1k & 10k & 100k & 1M \\
 &  & 6.27E-03 / 2.40 & 1.90E-02 / 12.9 & 1.62E-01 / 94.0 & 1.72 / 940  & 17.0 / 9375 \\
\cmidrule(l){2-7} 
 
\multirow{4}{*}{\makecell{\includegraphics[scale=0.5,trim={0cm 0.11cm 0cm 0cm},clip]{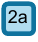} \texttt{gate-cluster} \\ O($s*\log(s) + s + \frac{m}{s} * \log(\frac{m}{s})$)}} & \multirow{2}{*}{\makecell{Num Domains \\ ($10$ Clusters)}} & 100 & 1k & 10k & 100k & 1M \\
 &  & 1.24E-03 / 2.17 & 6.43E-03 / 11.6 & 3.16E-02 / 99.5 & 3.06E-01 / 962 & 3.10 / 9561 \\ 
 \cmidrule(l){3-7}
 
 & \multirow{2}{*}{\makecell{Num Clusters \\ (1M Domains)}} & 10 & 100 & 1k & 10k & 100k \\
 &  & 1.85 / 9610 & 1.72E-01 / 9162 & 1.45E-01 / 9349 & 1.78 / 9674 & 22.8 / 9803 \\ \bottomrule 
\end{tabular}
\caption{\small The run-time and memory usage of each component in our scheme with varying parameter values. We change one parameter at a time while keeping the other parameters constant. The constant parameter is shown inside a parenthesis. The theoretical complexity is shown in the first column with $n$ referring to the number of tokens being evaluated. For descriptions of the other variables, refer to Table~\ref{tab:hyperparam}.}
\label{tab:runtime}
\vspace{-0.2in}
\end{table*}

\noindent\textbf{Forward Pass.} 
Table~\ref{tab:runtime} shows the breakdown of the run-time and memory usage for different model sizes and use cases. The result suggests that the largest contribution to the overall latency comes from the forward pass on the model. It is therefore important that in the case of output aggregation, we run the forward pass on each of the $k$ experts in parallel to reduce the latency. This issue would not be faced in parameter aggregation as the top-$k$ experts would be merged into one, resulting in a single forward pass. Both vectorizing and aggregation algorithms have comparatively small latencies.

\noindent\textbf{Gating Algorithms.} 
The table also shows the performance comparison between \texttt{gate-cluster} and \texttt{gate-pairwise}. As expected, the hierarchical search with clustering (\texttt{gate-cluster}) is faster than the pairwise search (\texttt{gate-pairwise}) for the same number of domains ($m$).
However, the run-time overhead of both gating algorithms is small compared to the forward pass latency. For security domains up to tens of thousands, \texttt{gate-pairwise} can be used with no noticeable impact on the end-to-end latency. 
For a very large number of security domains, \texttt{gate-cluster} provides a more scalable solution.


For \texttt{gate-cluster}, the table also shows that the number of clusters ($s$) and the number of domains ($m$) need to be balanced
for low latency. 
\texttt{gate-cluster} shows poor performance when the number of clusters is either too large or small 
because it performs tiered sorting; first sorting to find the best cluster, then sorting to find the top-$k$ experts within the cluster.
With too many clusters, the search for the best cluster is slow. When there are only a small number of very large clusters, the search within a cluster takes long.
{\small
\begin{equation}
\begin{gathered}
\text{top}(k) = \text{gate}(m) + \text{vectorize}(c) \\
    \text{time} = \text{top}(k) + \max 
\Bigg\{
\begin{array}{c}
      (\text{forward-pass}() + \text{aggregate}()) \times r \\
      \text{top}(k)
\end{array}
\Bigg\} \times \frac{n}{r}
\end{gathered}
\label{eq:runtime}
\end{equation}
}%

\begin{table}
\centering
\footnotesize
\begin{tabular}{@{}|c|ccc|@{}}
\toprule
\multirow{2}{*}{Use Case} & \multicolumn{3}{c|}{Latency (s) / Memory (MiB)} \\ \cmidrule{2-4}
& Our Case  & Baseline & Overhead \\ \midrule

\makecell{Next Word \\ Prediction (1)}  & 0.167 / 4051 & 0.166 / 3618  & 1.006 / 1.12  \\ 

\makecell{Next Phrase \\ Prediction (10)} & 0.358 / 4060 & 0.354 / 3627  & 1.011 / 1.12 \\ 

\makecell{Document \\ Paraphrasing (500)} & 10.6 / 3646  & 10.4 / 3213  & 1.019 / 1.13 \\ \bottomrule 

\end{tabular}
\caption{\small The average latency and memory usage to generate $1$, $10$, or $500$ tokens. The average latency is computed by generating $1$k tokens. The average memory usage is shown for one GPU (our scheme uses 3 GPUs in parallel for $k=3$). The experiments use GPT-2 Large and \texttt{gate-pairwise} with the following parameters: $m=10,000$, $k=3$, $r=200$, and $c=100$.}
\label{tab:end-to-end}
\vspace{-0.25in}
\end{table}

\noindent\textbf{End-to-End Performance.}
Given the run-time of each component in Table~\ref{tab:runtime}, the end-to-end latency can be calculated as shown in Equation~\ref{eq:runtime}.
Table~\ref{tab:end-to-end} shows the end-to-end latency of \texttt{gate-pairwise} with GPT-2 (large) model for three different use cases that generate a different number of tokens: next-word prediction (1 token), next-phrase prediction (10 tokens), and document paraphrasing ($500$ tokens). 
We show the performance of \texttt{gate-pairwise} as it represents the slowest gating algorithm. 
The average latency is obtained by generating 1,000 tokens for each task for a large number of security domains (10,000). 
The average memory usage per GPU is shown, since the memory capacity of each GPU is often the limiting factor in running a large language model. 
The results show that the overall overhead of the proposed parametric IFC scheme is reasonable even for a large number of security domains; the worst-case latency overhead is only 1.9\%, and the worst-case memory usage overhead is 13\%. 

%% file: ablation.tex
\vspace{-0.1in}
\subsection{Ablation Study: Activated Experts}
\label{sec:ablation}
\vspace{-0.05in}

\begin{figure}
    \centering\includegraphics{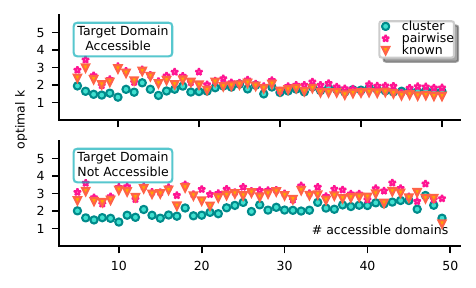}
    \vspace{-0.2in}
    \caption{\small The optimal $k$ value as a function of the number of accessible domains for $150$ access policies (top graph). Each data point is the geometric mean over $3 \times 50$ (access policy, domain) combinations. The bottom graph shows a sub-sample of the $150$ access policies where the test domain is inaccessible.} 
    \vspace{-0.3in}
    \label{fig:ablate-k-pol}
\end{figure}

Here, we study how sensitive our scheme is to the number of activated experts. A study of some of the other parameters in Table~\ref{tab:hyperparam} is provided in Appendix~\ref{sec:appendix-ablation}. In the following experiments, we run our schemes with $150$ different access policies on the validation sets of all $50$ domains of the Pushshift dataset. 


Figure~\ref{fig:ablate-k-pol} shows how the number of optimal experts ($k$) changes with the number of accessible domains for $\approx150$ access policies. The bottom graph in Figure~\ref{fig:ablate-k-pol} demonstrates the trend only for access policies where the target domain is not accessible. In the top graph, the result indicates that using more experts (larger $k$) is beneficial when only a very small number of domains are accessible. When only a few domains are accessible, it is unlikely that any of the accessible domains are directly related to the text being evaluated, and hence it is useful to have multiple relevant experts. On the other hand, as more domains become accessible, it is more likely that one of the accessible experts is trained with the data close to the input text, and can already provide high accuracy without other experts. 
Finally, in the bottom graph, when the input's domain needs to be estimated or is not part of the accessible experts, then $2-4$ experts are needed for the best accuracy. In particular, this scenario benefits from more experts overall---which is expected since no single expert is ``best'' for the test domain. 





%% file: limitations.tex
\vspace{-0.2in}
\subsection{Limitations}
\label{sec:limitations}
\vspace{-0.05in}
While our approach provides strict security guarantees, there are a few limitations. First, frequent and unpredictable input domain shifts from the user can lead to sub-optimal gating (and lower accuracy) in our approach when the input domain label is not provided.
Second, output ensembling uses k-times as many resources, each to run the k experts. While this may be solved by parameter merging, our experiments show that parameter merging does not perform as well as output ensembling.
Future work is needed to address these limitations.

%% file: extension.tex
\vspace{-0.1in}
\section{Future Work: Quantitative IFC}
\label{sec:extension}
\vspace{-0.1in}


The strict non-interference that we use in this paper can be generalized to allow a limited amount of information leakage from inaccessible training data to a model output. Doing so can improve inference accuracy for queries that are out-of-domain for a given access policy.
We generalize the definition of non-interference to $\epsilon$-non-interference ($\epsilon$-NI) below, which limits information leakage from inaccessible domains in $D - a_i$ to $\epsilon>0$ in a similar manner to differential privacy~\cite{dwork2014algorithmic}.


\noindent {\textbf{$\epsilon$-Non-Interference.}} Let the dataset $D$, access policy $a_i$ and relation $=_{a_i}$ be as defined in Section \ref{sec:ni}.
An inference algorithm $\mc{M}$ satisfies $\epsilon$-Non-Interference ($\epsilon$-NI) with respect to access policy $a_i$ if for all queries $\pmb{x}$ and all output sets $\mc{O}$ the following holds:
\begin{align*}
    \forall D, D^{\prime}: D =_{a_i} D^{\prime} \wedge \\
    \mc{M_D}(a_i, \pmb{x}) &\Downarrow o \wedge \\
    \mc{M_D^{\prime}}(a_i, \pmb{x}) &\Downarrow o^{\prime} \Rightarrow \\
    e^{-\epsilon} \leq \frac{P(o \in \mc{O})}{P(o^{\prime} \in \mc{O})} &\leq e^{\epsilon} 
\end{align*}

Our definition of $\epsilon$-NI is closely related to that of private prediction~\cite{dwork2018privacy}.
However, one key difference is that the notion of adjacency is not with respect to single training examples, but of entire sets of domains. For instance, $D$ is adjacent to all datasets $D'$ (i.e., $D =_{a_i} D'$) where data from the accessible domains $j : j \in a_i$ are the same. The notion of adjacency used in subsample-and-aggregate style protocols such as PATE~\cite{papernot2017semisupervised} and SubMix~\cite{ginart2022submix} can be viewed as a special case of our definition, restricted to singleton inaccessible domains.

In general, $\epsilon$-NI offers more flexibility than differential privacy due to the fact that the access policy only considers predetermined subsets of data (i.e., domains) that may be known ahead of time. It may be possible to design specialized inference mechanisms that satisfy $\epsilon$-NI while providing high utility. We leave the exploration and design of such mechanisms for future work.



%% file: related-works.tex
\vspace{-0.1in}
\section{Related Work}
\vspace{-0.1in}

\noindent \textbf{Mixture of Experts.} 
Mixture-of-expert (MoE) language models have been studied extensively, especially in the context of efficient model scaling. These models also tend to employ gating and ensembling, however their methods do not enforce the strict isolation that is needed for Parametric IFC. Chen et al.~\cite{chen2022mod} propose to optimize multi-task learning through co-operation and specialization between different modular experts in an MoE setting. 
Tang et al.~\cite{tang2002input} study the significance of clustering as a pre-processing step to aid routing inputs to specific experts. Work has also been done in developing MoE schemes with a large number of experts~\cite{shazeer2017outrageously,lepikhin2021gshard}, but using learned routing.
There have also been efforts in controlled generation~\cite{dathathri2020plug,keskar2019ctrl,liu2021-dexperts}---i.e., controlling various attributes (such as sentiment) of the generated text, as opposed to blind generation that most LLMs support today. 

\noindent \textbf{Retrieval Augmented Generation}
There have been several prior works exploring retrieval augmented generation, especially in the case of knowledge intensive tasks like question answering~\cite{guu2020retrieval,lewis2020retrieval,borgeaud2022improving,khandelwal2019generalization}. However, IFC in ML is a new area of research with very limited prior literature. There has been concurrent work in retrieval-based IFC~\cite{wutschitz2023rethinking}, although to the best of our knowledge, we are still the first to define and propose IFC for parametric models, which we believe is a critical missing component in enabling IFC for full ML models.


\noindent \textbf{Machine Unlearning.}
Another related area in ML is machine unlearning, which seeks to remove the traces of certain data from a trained model. One approach is to apply a model update that removes the influence of the sample to be unlearned~\cite{cao2015towards, ginart2019making, guo2020certified, sekhari2021remember}. Another solution uses a sharding+ensembling based mechanism for unlearning~\cite{bourtoule2021machine}. 
While related, machine unlearning is different from information flow control; machine unlearning aims to handle infrequent removal of an arbitrary subset of training data whereas information flow control aims to control the influence of training data for each query using security domains defined at training time.
When coarse-grained unlearning is acceptable, the modular architecture that we propose can also be useful for unlearning as it supports adding or removing security domains after training.

\noindent \textbf{Federated Learning:} 
Federated Learning (FL)~\cite{mcmahan2017communication} trains a model with data disaggregated across multiple parties. In FL, each party trains a model locally using its private data and sends it to an aggregator, which aggregates the model to produce a generalized model.
%
While FL eliminates the need to collect raw data at a centralized location during training, it still produces a single model that contains information from all the training data~\cite{mcmahan_dp_fl, sanjay_cpa}, requiring further protection if the information leakage must be limited~\cite{mcmahan_dp_fl, gboard_fl_dp1, gboard_fl_dp2}.
%
It would be interesting future work to complement FL with IFC by using FL to train experts for individual security domains. Such a hybrid design has the potential to provide unique privacy-enhancing properties both during the training of experts (through FL) and inference (through IFC).
%

\noindent \textbf{Differential Privacy:}
Differential Privacy (DP) can be applied to ML training~\cite{abadi2016deep} and inference~\cite{dwork2018privacy} to prevent training data from leaking through the trained model~\cite{balle2022reconstructing, guo2022bounding, guo2023analyzing}. In particular, recent works adopted differentially private training methods~\cite{li2021large,bu2024automatic,zhao2022provably} and inference methods~\cite{ginart2022submix} for large language models to protect training data.
As discussed in Section~\ref{sec:dp}, DP training provides an orthogonal protection to IFC, and the two are not directly comparable.

%% file: conclusion.tex
\vspace{-0.1in}
\section{Conclusion}
\vspace{-0.1in}

In this work, we present the concept of Information Flow Control (IFC) in machine learning, including a mathematical formulation of the problem. We then propose a method to provide IFC for parametric ML models and show how the IFC capability can be supported through modest modifications for a Transformer-based language model. 
The evaluation results based on the datasets from Pushshift.io and Codeparrot show that the proposed parametric IFC architecture can enable strong control of information flow from training data to model outputs with low overhead.

\vspace{-0.1in}
\section{Acknowledgments}
\vspace{-0.1in}
This work is partly supported by the U.S. National Science Foundation under award No. CNS-2211599,
a research internship at Meta, and a fellowship from Cisco. 
Any opinions, findings and conclusions or recommendations expressed in this material are those 
of the author(s) and do not necessarily reflect the views of the U.S. National Science Foundation, Meta, or Cisco.
The authors would also like to thank Drew Zagieboylo and the anonymous reviewers for their constructive comments and help with the IFC formalism.
\vspace{-0.1in}

%% file: appendix.tex
\section{Dataset}
\begin{figure*} 
\begin{center}
\includegraphics[trim={0.2cm 0 0.2cm 0},clip]{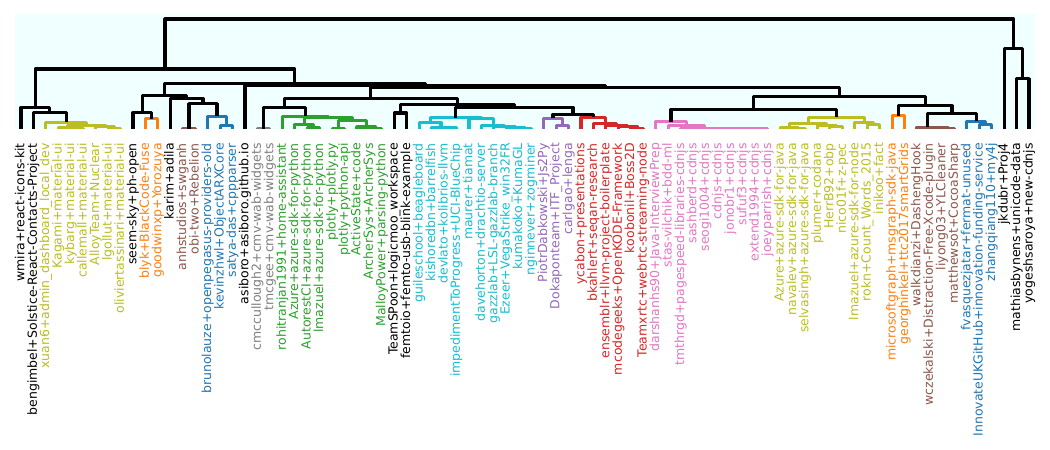}
\vspace{-0.2in}
\caption{\small A dendogram of the Codeparrot dataset representing how closely related each of the security domains are to each other. Furthermore, the color codes represent the clusters that these domains are divided into for \texttt{gate-cluster}.} \label{fig:github-dendogram}
\vspace{-0.3in}
\end{center}
\end{figure*}

\begin{figure} 
\begin{center}
\includegraphics[trim={0.18cm 0 0.2cm 0},clip]{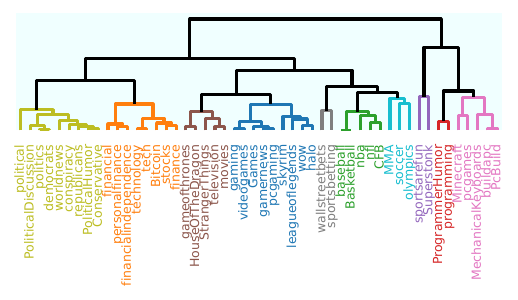}
\caption{\small A dendogram representing how closely related each of the $50$ domains from our Pushshift dataset are to each other. Furthermore, the color codes represent the clusters that these domains are divided into for  \texttt{gate-cluster}.} \label{fig:pushshift-dendogram}
\end{center}
\vspace{-0.3in}
\end{figure}

\vspace{-0.1in}
\label{sec:appendix-dataset}
Figures~\ref{fig:github-dendogram} and \ref{fig:pushshift-dendogram} show a relationship dendogram of the Codeparrot and Pushshift domains, showing how closely related different domains are. The color coding indicates the clusters that we use for {\tt gate-cluster}, implying that our clustering algorithm naturally clusters closely related domains. 



\vspace{-0.1in}
\section{Ablation Study (cont'd)}
\label{sec:appendix-ablation}
\vspace{-0.1in}



\subsection{Sample Text Length}
\begin{figure}
    \centering
    \includegraphics{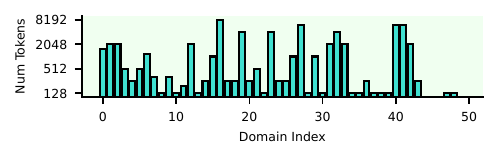}
    \vspace{-0.35in}
    \caption{\small The minimum number of tokens needed to correctly identify each of the $50$ domains in the Pushshift dataset. Domains are sorted from largest to smallest.}
    \label{fig:ablate-context}
\end{figure}

\vspace{-0.1in}
Figure~\ref{fig:ablate-context} shows the number of tokens in the sample text that is needed to correctly identify each of the $50$ domains in the Pushshift dataset. The result shows that a fair number of domains can be identified correctly with as few as $128$ tokens. However, there are a few domains that need as many as $~8$k tokens to be identified, indicating that there are multiple domains that are quite similar. It is also interesting to note that there is no clear correlation between the number of tokens needed to identify a domain and the size of the domain. 


\vspace{-0.1in}
\subsection{All Accessible Experts vs Top-K}
\begin{figure}
    \centering
    \includegraphics{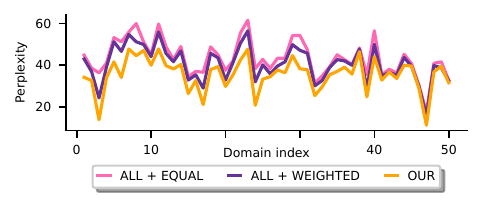}
    \vspace{-0.1in}
    \caption{\small Evaluating all $50$ Pushshift domains with our top-k scheme and comparing it to two naive algorithm of turning on all $50$ experts with a) equal weighting (ALL + EQUAL), b) with our weighing scheme (ALL + WEIGHTED). We see that choosing only the top-K experts and weighing them intelligently (OUR) performs superior to both.}
    \label{fig:all-top-k}
    \vspace{-0.2in}
\end{figure}

\vspace{-0.05in}
We evaluate two naive schemes of turning on all accessible experts and compare it with our scheme in Figure~\ref{fig:all-top-k}. Here, we assume that all $50$ experts in the Pushshift dataset are accessible, and thus ensemble all $50$ experts to evaluate each of the test datasets. We see that giving equal weights to all $50$ experts performs the worst. Intelligently weighing all $50$ experts based on their relevance to their test dataset performs better, however, the best performance comes from activating just the top-K out of all $50$ experts and weighing the K experts based on their relevance, which is the technique employed in our IFC framework.

%% file: main.bbl
\begin{thebibliography}{10}

\bibitem{abadi2016deep}
M.~Abadi, A.~Chu, I.~Goodfellow, H.~B. McMahan, I.~Mironov, K.~Talwar, and
  L.~Zhang.
\newblock Deep learning with differential privacy.
\newblock In {\em ACM SIGSAC Conference on Computer and Communications Security
  (CCS)}, 2016.

\bibitem{balle2022reconstructing}
B.~Balle, G.~Cherubin, and J.~Hayes.
\newblock Reconstructing training data with informed adversaries.
\newblock In {\em IEEE Symposium on Security and Privacy (SP)}, 2022.

\bibitem{baumgartner2020pushshift}
J.~Baumgartner, S.~Zannettou, B.~Keegan, M.~Squire, and J.~Blackburn.
\newblock The {P}ushshift {R}eddit dataset.
\newblock In {\em International AAAI Conference on Web and Social Media}, 2020.

\bibitem{black2022gpt}
S.~Black, S.~Biderman, E.~Hallahan, Q.~Anthony, L.~Gao, L.~Golding, H.~He,
  C.~Leahy, K.~McDonell, J.~Phang, et~al.
\newblock {GPT}-{N}eo{X}-20{B}: An open-source autoregressive language model.
\newblock {\em arXiv preprint arXiv:2204.06745}, 2022.

\bibitem{borgeaud2022improving}
S.~Borgeaud, A.~Mensch, J.~Hoffmann, T.~Cai, E.~Rutherford, K.~Millican, G.~B.
  Van Den~Driessche, J.-B. Lespiau, B.~Damoc, A.~Clark, et~al.
\newblock Improving language models by retrieving from trillions of tokens.
\newblock In {\em International Conference on Machine Learning (ICML)}, 2022.

\bibitem{bourtoule2021machine}
L.~Bourtoule, V.~Chandrasekaran, C.~A. Choquette-Choo, H.~Jia, A.~Travers,
  B.~Zhang, D.~Lie, and N.~Papernot.
\newblock Machine unlearning.
\newblock In {\em IEEE Symposium on Security and Privacy (SP)}, 2021.

\bibitem{bu2024automatic}
Z.~Bu, Y.-X. Wang, S.~Zha, and G.~Karypis.
\newblock Automatic clipping: Differentially private deep learning made easier
  and stronger.
\newblock {\em Advances in Neural Information Processing Systems (NeurIPS)},
  2024.

\bibitem{cao2015towards}
Y.~Cao and J.~Yang.
\newblock Towards making systems forget with machine unlearning.
\newblock In {\em IEEE Symposium on Security and Privacy (SP)}, 2015.

\bibitem{carlini_diffusion}
N.~Carlini, J.~Hayes, M.~Nasr, M.~Jagielski, V.~Sehwag, F.~Tram{\`{e}}r,
  B.~Balle, D.~Ippolito, and E.~Wallace.
\newblock Extracting training data from diffusion models.
\newblock In {\em 32nd USENIX Security Symposium}, 2023.

\bibitem{carlini2021extracting}
N.~Carlini, F.~Tramer, E.~Wallace, M.~Jagielski, A.~Herbert-Voss, K.~Lee,
  A.~Roberts, T.~B. Brown, D.~Song, U.~Erlingsson, et~al.
\newblock Extracting training data from large language models.
\newblock In {\em USENIX Security Symposium}, 2021.

\bibitem{chen2022mod}
Z.~Chen, Y.~Shen, M.~Ding, Z.~Chen, H.~Zhao, E.~Learned-Miller, and C.~Gan.
\newblock Mod-{S}quad: Designing mixture of experts as modular multi-task
  learners.
\newblock {\em arXiv preprint arXiv:2212.08066}, 2022.

\bibitem{codeparrot}
Codeparrot.
\newblock Github-code by codeparrot, 2022.

\bibitem{dathathri2020plug}
S.~Dathathri, A.~Madotto, J.~Lan, J.~Hung, E.~Frank, P.~Molino, J.~Yosinski,
  and R.~Liu.
\newblock Plug and play language models: a simple approach to controlled text
  generation.
\newblock In {\em International Conference on Learning Representations (ICLR)},
  2020.

\bibitem{devlin2018bert}
J.~Devlin, M.-W. Chang, K.~Lee, and K.~Toutanova.
\newblock {BERT}: Pre-training of deep bidirectional transformers for language
  understanding.
\newblock {\em arXiv preprint arXiv:1810.04805}, 2018.

\bibitem{dwork2018privacy}
C.~Dwork and V.~Feldman.
\newblock Privacy-preserving prediction.
\newblock In {\em Conference On Learning Theory}, 2018.

\bibitem{dwork2006calibrating}
C.~Dwork, F.~McSherry, K.~Nissim, and A.~Smith.
\newblock Calibrating noise to sensitivity in private data analysis.
\newblock In {\em Theory of Cryptography: Third Theory of Cryptography
  Conference}, 2006.

\bibitem{dwork2014algorithmic}
C.~Dwork, A.~Roth, et~al.
\newblock The algorithmic foundations of differential privacy.
\newblock {\em Foundations and Trends in Theoretical Computer Science}, 2014.

\bibitem{fan2019eli5}
A.~Fan, Y.~Jernite, E.~Perez, D.~Grangier, J.~Weston, and M.~Auli.
\newblock Eli5: Long form question answering.
\newblock {\em arXiv preprint arXiv:1907.09190}, 2019.

\bibitem{ginart2019making}
A.~Ginart, M.~Guan, G.~Valiant, and J.~Y. Zou.
\newblock Making ai forget you: Data deletion in machine learning.
\newblock {\em Advances in Neural Information Processing Systems (NeurIPS)},
  2019.

\bibitem{ginart2022submix}
A.~Ginart, L.~van~der Maaten, J.~Zou, and C.~Guo.
\newblock Submix: Practical private prediction for large-scale language models.
\newblock {\em arXiv preprint arXiv:2201.00971}, 2022.

\bibitem{copilotapi}
Github.
\newblock About github copilot, 2024.

\bibitem{goguen1982security}
J.~A. Goguen and J.~Meseguer.
\newblock Security policies and security models.
\newblock In {\em IEEE Symposium on Security and Privacy (SP)}, 1982.

\bibitem{guo2020certified}
C.~Guo, T.~Goldstein, A.~Hannun, and L.~Van Der~Maaten.
\newblock Certified data removal from machine learning models.
\newblock In {\em 37th International Conference on Machine Learning (ICML)},
  2020.

\bibitem{guo2022bounding}
C.~Guo, B.~Karrer, K.~Chaudhuri, and L.~van~der Maaten.
\newblock Bounding training data reconstruction in private (deep) learning.
\newblock In {\em International Conference on Machine Learning (ICML)}, 2022.

\bibitem{guo2023analyzing}
C.~Guo, A.~Sablayrolles, and M.~Sanjabi.
\newblock Analyzing privacy leakage in machine learning via multiple hypothesis
  testing: A lesson from {F}ano.
\newblock In {\em International Conference on Machine Learning (ICML)}, 2023.

\bibitem{embeddingShortcomings}
M.~Gupta.
\newblock Text vectorization algorithms in {NLP}, 2022.

\bibitem{gururangan2021demix}
S.~Gururangan, M.~Lewis, A.~Holtzman, N.~A. Smith, and L.~Zettlemoyer.
\newblock {DEM}ix layers: Disentangling domains for modular language modeling.
\newblock {\em arXiv preprint arXiv:2108.05036}, 2021.

\bibitem{guu2020retrieval}
K.~Guu, K.~Lee, Z.~Tung, P.~Pasupat, and M.~Chang.
\newblock Retrieval augmented language model pre-training.
\newblock In {\em International Conference on Machine Learning (ICML)}, 2020.

\bibitem{houlsby2019parameter}
N.~Houlsby, A.~Giurgiu, S.~Jastrzebski, B.~Morrone, Q.~De~Laroussilhe,
  A.~Gesmundo, M.~Attariyan, and S.~Gelly.
\newblock Parameter-efficient transfer learning for {NLP}.
\newblock In {\em International Conference on Machine Learning}, 2019.

\bibitem{hu2021lora}
E.~J. Hu, P.~Wallis, Z.~Allen-Zhu, Y.~Li, S.~Wang, L.~Wang, W.~Chen, et~al.
\newblock {L}o{RA}: Low-rank adaptation of large language models.
\newblock In {\em International Conference on Learning Representations (ICLR)},
  2021.

\bibitem{izacard2022few}
G.~Izacard, P.~Lewis, M.~Lomeli, L.~Hosseini, F.~Petroni, T.~Schick,
  J.~Dwivedi-Yu, A.~Joulin, S.~Riedel, and E.~Grave.
\newblock Few-shot learning with retrieval augmented language models.
\newblock {\em arXiv preprint arXiv:2208.03299}, 2022.

\bibitem{izmailov2018averaging}
P.~Izmailov, D.~Podoprikhin, T.~Garipov, D.~Vetrov, and A.~G. Wilson.
\newblock Averaging weights leads to wider optima and better generalization.
\newblock In {\em Uncertainty in Artificial Intelligence 2018 (UAI)}, 2018.

\bibitem{bagofwords}
V.~Jayaswal.
\newblock Text vectorization: Bag of {W}ords ({BOW}).
\newblock {\em Medium, Towards Data Science}, 2020.

\bibitem{sanjay_cpa}
S.~Kariyappa, C.~Guo, K.~Maeng, W.~Xiong, G.~E. Suh, M.~K. Qureshi, and
  H.-H.~S. Lee.
\newblock Cocktail party attack: Breaking aggregation-based privacy in
  federated learning using independent component analysis.
\newblock In {\em International Conference on Machine Learning (ICML)}, 2023.

\bibitem{keskar2019ctrl}
N.~S. Keskar, B.~McCann, L.~R. Varshney, C.~Xiong, and R.~Socher.
\newblock {CTRL}: A conditional transformer language model for controllable
  generation.
\newblock {\em arXiv preprint arXiv:1909.05858}, 2019.

\bibitem{khandelwal2019generalization}
U.~Khandelwal, O.~Levy, D.~Jurafsky, L.~Zettlemoyer, and M.~Lewis.
\newblock Generalization through memorization: Nearest neighbor language
  models.
\newblock {\em arXiv preprint arXiv:1911.00172}, 2019.

\bibitem{kifer_no_free_lunch}
D.~Kifer and A.~Machanavajjhala.
\newblock No free lunch in data privacy.
\newblock In {\em ACM SIGMOD International Conference on Management of Data},
  2011.

\bibitem{lepikhin2021gshard}
D.~Lepikhin, H.~Lee, Y.~Xu, D.~Chen, O.~Firat, Y.~Huang, M.~Krikun, N.~Shazeer,
  and Z.~Chen.
\newblock {GS}hard: scaling giant models with conditional computation and
  automatic sharding.
\newblock In {\em International Conference on Learning Representations (ICLR)},
  2021.

\bibitem{lewis2020retrieval}
P.~Lewis, E.~Perez, A.~Piktus, F.~Petroni, V.~Karpukhin, N.~Goyal,
  H.~K{\"u}ttler, M.~Lewis, W.-t. Yih, T.~Rockt{\"a}schel, et~al.
\newblock Retrieval-augmented generation for knowledge-intensive {NLP} tasks.
\newblock {\em Advances in Neural Information Processing Systems (NeurIPS)},
  2020.

\bibitem{li2022branch}
M.~Li, S.~Gururangan, T.~Dettmers, M.~Lewis, T.~Althoff, N.~A. Smith, and
  L.~Zettlemoyer.
\newblock Branch-{T}rain-{M}erge: Embarrassingly parallel training of expert
  language models.
\newblock {\em arXiv preprint arXiv:2208.03306}, 2022.

\bibitem{li2021large}
X.~Li, F.~Tramer, P.~Liang, and T.~Hashimoto.
\newblock Large language models can be strong differentially private learners.
\newblock {\em arXiv preprint arXiv:2110.05679}, 2021.

\bibitem{liu2021-dexperts}
A.~Liu, M.~Sap, X.~Lu, S.~Swayamdipta, C.~Bhagavatula, N.~A. Smith, and
  Y.~Choi.
\newblock {DE}xperts: decoding-time controlled text generation with experts and
  anti-experts.
\newblock In {\em 59th Annual Meeting of the Association for Computational
  Linguistics (ACL) and the 11th International Joint Conference on Natural
  Language Processing}, 2021.

\bibitem{liu2019roberta}
Y.~Liu, M.~Ott, N.~Goyal, J.~Du, M.~Joshi, D.~Chen, O.~Levy, M.~Lewis,
  L.~Zettlemoyer, and V.~Stoyanov.
\newblock {R}o{BERT}a: A robustly optimized {BERT} pretraining approach.
\newblock {\em arXiv preprint arXiv:1907.11692}, 2019.

\bibitem{masoudnia2014mixture}
S.~Masoudnia and R.~Ebrahimpour.
\newblock Mixture of experts: a literature survey.
\newblock {\em The Artificial Intelligence Review}, 2014.

\bibitem{mcmahan2017communication}
B.~McMahan, E.~Moore, D.~Ramage, S.~Hampson, and B.~A. y~Arcas.
\newblock Communication-efficient learning of deep networks from decentralized
  data.
\newblock In {\em Artificial Intelligence and Statistics}, 2017.

\bibitem{mcmahan_dp_fl}
H.~B. McMahan, D.~Ramage, K.~Talwar, and L.~Zhang.
\newblock Learning differentially private recurrent language models.
\newblock In {\em International Conference on Learning Representations (ICLR)},
  2018.

\bibitem{ngram}
A.~Nair.
\newblock Leveraging {N}-grams to extract context from text.
\newblock {\em Medium, Towards Data Science}, 2021.

\bibitem{carlini_chatgpt}
M.~Nasr, N.~Carlini, J.~Hayase, M.~Jagielski, A.~F. Cooper, D.~Ippolito, C.~A.
  Choquette-Choo, E.~Wallace, F.~Tram{\`e}r, and K.~Lee.
\newblock Scalable extraction of training data from (production) language
  models.
\newblock {\em arXiv preprint arXiv:2311.17035}, 2023.

\bibitem{papernot2017semisupervised}
N.~Papernot, M.~Abadi, {\'U}.~Erlingsson, I.~Goodfellow, and K.~Talwar.
\newblock Semi-supervised knowledge transfer for deep learning from private
  training data.
\newblock In {\em International Conference on Learning Representations (ICLR)},
  2017.

\bibitem{radford2018improving}
A.~Radford, K.~Narasimhan, T.~Salimans, I.~Sutskever, et~al.
\newblock Improving language understanding by generative pre-training.
\newblock {\em OpenAI}, 2018.

\bibitem{raffel2020exploring}
C.~Raffel, N.~Shazeer, A.~Roberts, K.~Lee, S.~Narang, M.~Matena, Y.~Zhou,
  W.~Li, and P.~J. Liu.
\newblock Exploring the limits of transfer learning with a unified text-to-text
  transformer.
\newblock {\em The Journal of Machine Learning Research (JMLR)}, 2020.

\bibitem{sekhari2021remember}
A.~Sekhari, J.~Acharya, G.~Kamath, and A.~T. Suresh.
\newblock Remember what you want to forget: Algorithms for machine unlearning.
\newblock {\em Advances in Neural Information Processing Systems (NeurIPS)},
  2021.

\bibitem{shazeer2017outrageously}
N.~Shazeer, A.~Mirhoseini, K.~Maziarz, A.~Davis, Q.~Le, G.~Hinton, and J.~Dean.
\newblock Outrageously large neural networks: The sparsely-gated
  mixture-of-experts layer.
\newblock In {\em International Conference on Learning Representations (ICLR)},
  2017.

\bibitem{shi2023large}
F.~Shi, X.~Chen, K.~Misra, N.~Scales, D.~Dohan, E.~H. Chi, N.~Sch{\"a}rli, and
  D.~Zhou.
\newblock Large language models can be easily distracted by irrelevant context.
\newblock In {\em International Conference on Machine Learning (ICML)}, 2023.

\bibitem{shi2022just}
W.~Shi, R.~Shea, S.~Chen, C.~Zhang, R.~Jia, and Z.~Yu.
\newblock Just fine-tune twice: selective differential privacy for large
  language models.
\newblock In {\em Empirical Methods in Natural Language Processing (EMNLP)},
  2022.

\bibitem{tang2002input}
B.~Tang, M.~I. Heywood, and M.~Shepherd.
\newblock Input partitioning to mixture of experts.
\newblock In {\em International Joint Conference on Neural Networks}, 2002.

\bibitem{vaswani2017attention}
A.~Vaswani, N.~Shazeer, N.~Parmar, J.~Uszkoreit, L.~Jones, A.~N. Gomez,
  {\L}.~Kaiser, and I.~Polosukhin.
\newblock Attention is all you need.
\newblock {\em Advances in Neural Information Processing Systems (NeurIPS)},
  2017.

\bibitem{wortsman2022model}
M.~Wortsman, G.~Ilharco, S.~Y. Gadre, R.~Roelofs, R.~Gontijo-Lopes, A.~S.
  Morcos, H.~Namkoong, A.~Farhadi, Y.~Carmon, S.~Kornblith, et~al.
\newblock Model soups: Averaging weights of multiple fine-tuned models improves
  accuracy without increasing inference time.
\newblock In {\em International Conference on Machine Learning (ICML)}, 2022.

\bibitem{wutschitz2023rethinking}
L.~Wutschitz, B.~K{\"o}pf, A.~Paverd, S.~Rajmohan, A.~Salem, S.~Tople,
  S.~Zanella-B{\'e}guelin, M.~Xia, and V.~R{\"u}hle.
\newblock Rethinking privacy in machine learning pipelines from an information
  flow control perspective.
\newblock {\em arXiv preprint arXiv:2311.15792}, 2023.

\bibitem{xu2023recomp}
F.~Xu, W.~Shi, and E.~Choi.
\newblock {RECOMP}: Improving retrieval-augmented {LM}s with compression and
  selective augmentation.
\newblock {\em arXiv preprint arXiv:2310.04408}, 2023.

\bibitem{gboard_fl_dp1}
Z.~Xu, Y.~Zhang, G.~Andrew, C.~A. Choquette{-}Choo, P.~Kairouz, H.~B. McMahan,
  J.~Rosenstock, and Y.~Zhang.
\newblock Federated learning of {G}board language models with differential
  privacy.
\newblock In {\em 61st Annual Meeting of the Association for Computational
  Linguistics (ACL): Industry Track}, 2023.

\bibitem{yang2019xlnet}
Z.~Yang, Z.~Dai, Y.~Yang, J.~Carbonell, R.~R. Salakhutdinov, and Q.~V. Le.
\newblock {XLN}et: Generalized autoregressive pretraining for language
  understanding.
\newblock {\em Advances in Neural Information Processing Systems (NeurIPS)},
  2019.

\bibitem{zhang2022opt}
S.~Zhang, S.~Roller, N.~Goyal, M.~Artetxe, M.~Chen, S.~Chen, C.~Dewan, M.~Diab,
  X.~Li, X.~V. Lin, et~al.
\newblock {OPT}: Open pre-trained transformer language models.
\newblock {\em arXiv preprint arXiv:2205.01068}, 2022.

\bibitem{gboard_fl_dp2}
Y.~Zhang, D.~Ramage, Z.~Xu, Y.~Zhang, S.~Zhai, and P.~Kairouz.
\newblock Private federated learning in {G}board.
\newblock {\em arXiv preprint arXiv:2306.14793}, 2023.

\bibitem{zhao2022provably}
X.~Zhao, L.~Li, and Y.-X. Wang.
\newblock Provably confidential language modelling.
\newblock In {\em North American Chapter of the Association for Computational
  Linguistics (ACL): Human Language Technologies}, 2022.

\end{thebibliography}
